\documentclass[10pt,twocolumn,letterpaper]{article}

\usepackage[pagenumbers]{cvpr} 

\usepackage[table]{xcolor}

\usepackage{amsmath,amsfonts,amssymb}
\usepackage{amsmath} 
\usepackage{amsfonts} 
\usepackage{amssymb} 
\usepackage{bm}
\usepackage{algorithm}
\usepackage{algpseudocode}
\usepackage{threeparttable} 
\usepackage{subcaption}

\newtheorem{prop}{Proposition}

\usepackage{todonotes}

\definecolor{cvprblue}{rgb}{0.21,0.49,0.74}
\usepackage[pagebackref,breaklinks,colorlinks,citecolor=cvprblue]{hyperref}

\def\paperID{16721} 
\def\confName{CVPR}
\def\confYear{2024}

\title{\vspace*{-10pt}Inversion-Free Image Editing with Natural Language \vspace*{-10pt}} 

\author{
    Sihan Xu\textsuperscript{\rm 1}\thanks{Authors contributed equally to this work.} \quad
    Yidong Huang\textsuperscript{\rm 1}\footnotemark[1] \quad
    Jiayi Pan\textsuperscript{\rm 2}\thanks{Work done while the author was at the University of Michigan.} \quad
    Ziqiao Ma\textsuperscript{\rm 1} \quad
    Joyce Chai\textsuperscript{\rm 1}
    \\
    \textsuperscript{\rm 1}University of Michigan \quad
    \textsuperscript{\rm 2}University of California, Berkeley 
    \\
    \url{https://sled-group.github.io/InfEdit/}
}

\begin{document}

\makeatletter
\let\@oldmaketitle\@maketitle
    \renewcommand{\@maketitle}{\@oldmaketitle
    \vspace*{-20pt}
    \centering
    \includegraphics[width=0.98\textwidth]{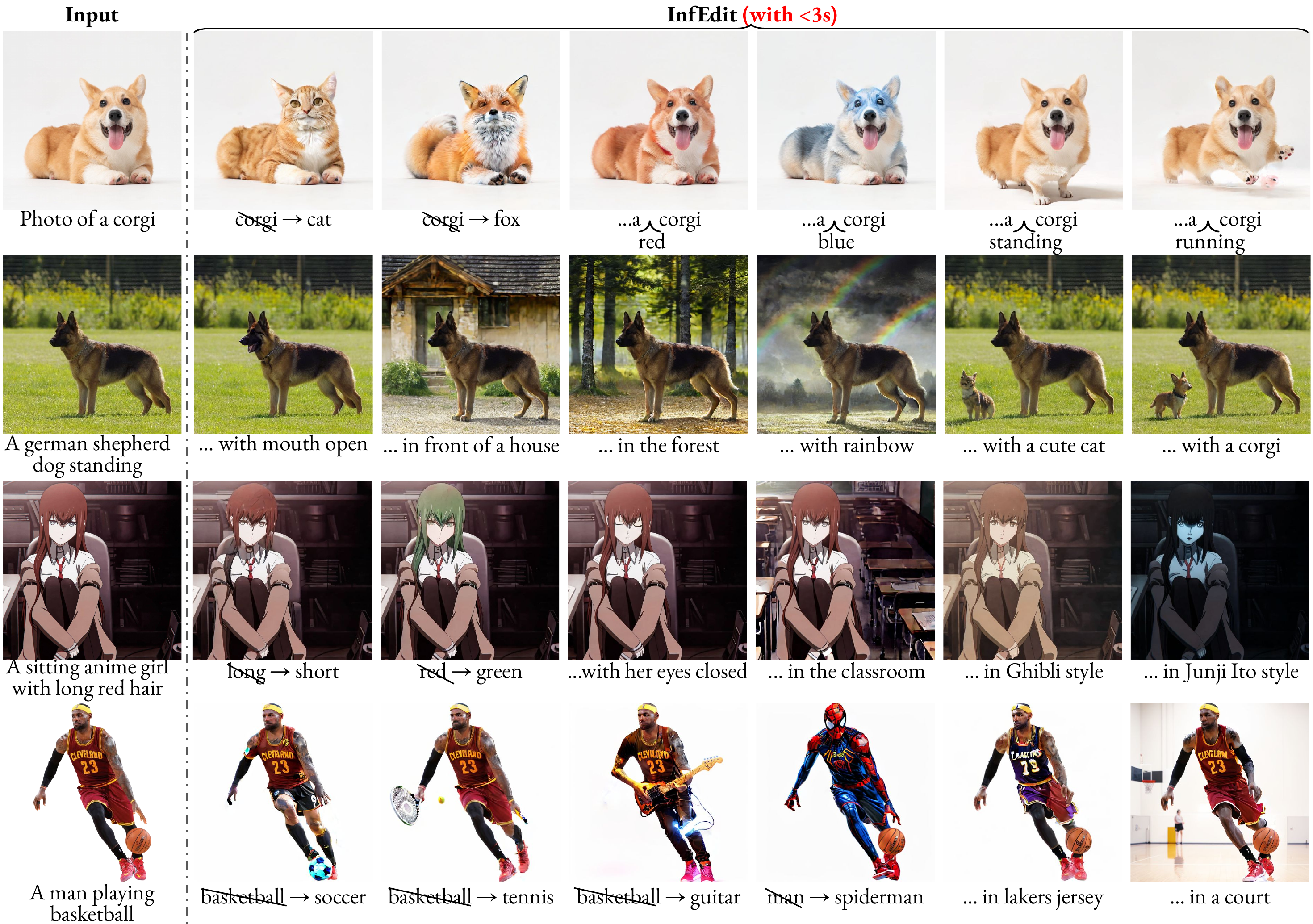}
    \vspace*{-10pt}
    \captionof{figure}{Our inversion-free editing (InfEdit) method demonstrates strong performance in various complex image editing tasks. \vspace*{-5pt}}
    \label{fig:teaser}
  \bigskip}
\makeatother

\maketitle

\begin{abstract}

\vspace*{-5pt}

Despite recent advances in inversion-based editing, text-guided image manipulation remains challenging for diffusion models.
The primary bottlenecks include 1) the time-consuming nature of the inversion process; 2) the struggle to balance consistency with accuracy; 3) the lack of compatibility with efficient consistency sampling methods used in consistency models.
To address the above issues, we start by asking ourselves if the inversion process can be eliminated for editing.
We show that when the initial sample is known, a special variance schedule reduces the denoising step to the same form as the multi-step consistency sampling.
We name this Denoising Diffusion Consistent Model (DDCM), and note that it implies a virtual inversion strategy without explicit inversion in sampling. 
We further unify the attention control mechanisms in a tuning-free framework for text-guided editing.
Combining them, we present inversion-free editing (InfEdit), which allows for consistent and faithful editing for both rigid and non-rigid semantic changes, catering to intricate modifications without compromising on the image's integrity and explicit inversion. 
Through extensive experiments, InfEdit shows strong performance in various editing tasks and also maintains a seamless workflow (less than 3 seconds on one single A40), demonstrating the potential for real-time applications.

\end{abstract}    
\section{Introduction}
\label{sec:intro}

Recent progress in image synthesis has been mostly driven by the development of Diffusion Models (DMs)~\cite{ho2020ddpm, song2020ddim}, which have outperformed traditional Generative Adversarial Networks (GANs)~\cite{goodfellow2020generative} and Variational Autoencoders (VAEs)~\cite{kingma2014auto} in various applications. 
A key factor in the wide success of DMs is their ability to incorporate diverse conditions, such as text~\cite{Ramesh2022HierarchicalTI}, images~\cite{zhang2023adding,mou2023t2i,xu2023cyclenet}, and even tactile input~\cite{yang2023generating}. 
Building upon DMs, Consistency Models (CMs)~\cite{song2023consistency} address the efficiency bottleneck by directly mapping noised samples along a trajectory to the same initial, promising \textit{self-consistency}.

Enabling text-guided DMs for editing real images using natural language has presented significant challenges.
Early methods typically require additional mask layers~\citep{nichol2022glide,bar2022text2live,couairon2023diffedit,avrahami2023spatext} or training~\cite{choi2021ilvr,zhao2022egsde,xu2023cyclenet}, which constrain their potential zero-shot application.
Motivated by DDIM inversion~\cite{song2020ddim}, a prevailing paradigm of \textit{inversion-based} editing has been established. 
The predominating methods along this line adopt \textit{optimization-based inversion}~\cite{mokady2023nti,li2023stylediffusion,dong2023prompt} by aligning the forward source latents with the DDIM inversion trajectory.
To address the issues of efficiency bottlenecks and far-from-ideal consistency, \textit{dual-branch} methods~\cite{wu2023cyclediffusion,ju2023direct} have been introduced, which separate the source and target branches individually, and iteratively calibrate the trajectory of the target branch.
However, inversion-based editing methods still face limitations in real-time and real-world language-guided image editing. 
Firstly, they typically rely on a lengthy inversion process to acquire the inversion branch as a series of anchors. 
Secondly, striking a balance between consistency and faithfulness remains challenging, even with extensive optimization or ways of calibrating the target branch. 
Lastly, these methods rely on variations of diffusion sampling, which are not compatible with the efficient consistency sampling using CMs.

To address the above challenges, we start by asking ourselves if the inversion process is really required for editing.
We show that when the initial sample is known, there exists a special variance schedule such that the denoising step takes the same form as the multi-step consistency sampling.
We name this Denoising Diffusion Consistent Model (DDCM), and note that it implies a sampling strategy that eliminates the inversion process. 
We further present Unified Attention Control (UAC), a tuning-free method that unifies attention control mechanisms for text-guided editing.
Combining them, we present an inversion-free editing (InfEdit) framework that allows for consistent and faithful editing for both rigid and non-rigid semantic changes, catering to intricate modifications without compromising on the image's integrity and explicit inversion. 
Through experiments, Inf-Edit shows strong performance in various editing tasks and also maintains a seamless workflow (less than 3s on one A40), demonstrating the potential for real-time editing.

\section{Preliminaries}
\label{sec:background}

\begin{figure*}[!htp]
    \centering
    \vspace*{-0.25cm}
    \includegraphics[width=1.0\linewidth]{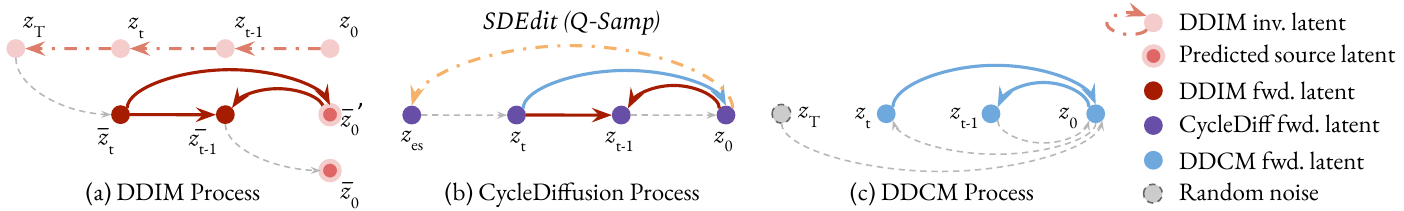}
    \vspace*{-20pt}
    \caption{While DDIM is prone to reconstruction error and requires iterative inversion, DDCM accepts any random noise to start with. It introduces a non-Markovian forward process in which $z_t$ directly points to the ground truth $z_0$ without neural prediction, and $z_{t-1}$ does not depend on the previous step $z_t$ like a consistency model. \vspace*{-5pt}}
    \label{fig::ddcm}
    \vspace*{-0.25cm}
\end{figure*}

\subsection{Diffusion Models}

Diffusion models (DMs)~\cite{ho2020ddpm} operate through a forward process that gradually adds Gaussian noises to data, described as follows:

\vspace*{-10pt}
\begin{equation}
\label{eq:zt}
z_{t} = \sqrt{{\alpha}_t} z_0 + \sqrt{1 - {\alpha}_t} \varepsilon\quad \varepsilon \sim \mathcal{N}(\bm{0},\bm{I})
\vspace*{-3pt}
\end{equation}
where $z_0$ is a sample from the data distribution, ${\alpha}_{1:T}$ specify a variance schedule for $t\sim [1, T]$.

The training objective involves a parameterized noise prediction network, $\varepsilon_{\theta}$, which aims to reverse the diffusion process. 
The training objective is to minimize the following loss based on a chosen metric function for measuring the distance between two samples $d(\cdot, \cdot)$:

\vspace*{-5pt}
\begin{equation}
\label{eq:obj}
\min_\theta \mathbb{E}_{z_0, \varepsilon, t} \big[d\left(\varepsilon, \varepsilon_{\theta}(z_t,t)\right)\big]
\vspace*{-3pt}
\end{equation}

Sampling from a diffusion model is an iterative process that progressively denoises the data. 
Following Eq (12) in~\citet{song2020ddim}, the denoising step at $t$ is formulated as:

\vspace*{-10pt}
\begin{equation}
\label{eq:fwd}
\begin{aligned}
    z_{t-1} = 
    & \sqrt{{{\alpha}}_{t-1}} \left( \frac{z_t - \sqrt{1 - {{\alpha}}_t} \varepsilon_{\theta}(z_t, t)}{\sqrt{{\alpha}}_t} \right) && \text{(predicted $z_0$)} \\
    & + \sqrt{1 - {\alpha}_{t-1} - \sigma_t^2} \cdot \varepsilon_{\theta}(z_t,t) && \text{(direction to $z_t$)} \\
    & + \sigma_t \varepsilon_t\quad \text{where } \varepsilon_t \sim \mathcal{N}(\bm{0},\bm{I}) && \text{(random noise)}
\end{aligned}
\end{equation}

DDPM sampling~\cite{ho2020ddpm} introduces a noise schedule $\sigma_t$ so that Eq (\ref{eq:fwd}) becomes Markovian.
By setting $\sigma_t$ to vanish, DDIM sampling~\cite{song2020ddim} results in an implicit probabilistic model with a deterministic forward process.

Following DDIM, we can use the function $f_\theta$ to predict and reconstruct $\bar{z_0}$ given $z_t$:

\vspace*{-15pt}
\begin{equation}
\label{eq:cond}
\bar{z}_0 
= f_\theta(z_t, t) 
= \left(z_t - \sqrt{1 - {\alpha}_t} \cdot \varepsilon_\theta(z_t, t) \right) / \sqrt{{\alpha}_t}
\vspace*{-3pt}
\end{equation}

Recently, Latent Diffusion Models (LDMs) \cite{rombach2022ldm} offer a new paradigm by operating in the latent space. 
The source latent $z_0$ is acquired by encoding a sample $x_0$ with an encoder $\mathcal{E}$, such that $z_0 = \mathcal{E}(x_0)$. 
So as to be reversed, the output can then be reconstructed by a decoder $\mathcal{D}$.
This framework presents a computationally efficient way to generate high-fidelity images, as the diffusion process is conducted in a latent space with lower dimensions.

\subsection{Consistency Models}

Consistency models (CMs)~\cite{song2023consistency} have recently been introduced, which greatly accelerate the generation process compared with previous DMs.
One notable property of CMs is \textit{self-consistency}, such that samples along a trajectory map to the sample initial. 
The key is a consistency function $f(z_t, t)$, which ensures a consistent distillation process by optimizing:

\vspace*{-15pt}
\begin{equation}
\label{eq:obj}
\min_{\theta,\theta^{-};\phi} \mathbb{E}_{z_0,t} \left[d\left(f_{\theta}(z_{t_{n+1}}, t_{n+1}), f_{\theta^{-}}(\hat{z}^{\phi}_{t_{n}}, t_{n})\right)\right]
\vspace*{-5pt}
\end{equation}
in which $f_{\theta}$ denotes a trainable neural network that parameterizes these consistent transitions, while $f_{\theta^{-}}$ represents a slowly updated target model used for consistency distillation, with the update rule $\theta^- \leftarrow \mu \theta ^{-} + (1-\mu) \theta$ given a decay rate $\mu$. 
The variable $\hat{z}^\phi_{t_n}$ denotes a one-step estimation of $z_{t_n}$ from $z_{t_{n+1}}$.

Sampling in CMs is carried out through a sequence of timesteps $\tau_{1:n} \in [t_0,T]$. 
Starting from an initial noise $\hat{z}_T$ and $z_0^{(T)} = f_\theta(\hat{z}_T, T)$, at each time-step $\tau_{i}$, the process samples $\varepsilon \sim \mathcal{N}(\bm{0}, \bm{I})$ and iteratively updates the \textit{Multistep Consistency Sampling} process:

\vspace*{-5pt}
\begin{equation}
\begin{aligned}
\hat{z}_{\tau_i} &= z_0^{(\tau_{i+1})} + \sqrt{\tau_i^2 - t_0^2 }\varepsilon \\
z_0^{(\tau_{i})} &= f_{\theta}(\hat{z}_{\tau_i}, \tau_i)
\end{aligned}
\end{equation}

Latent Consistency Models (LCMs) \cite{luo2023lcm} extend to accommodate a (text) condition $c$, which is crucial for text-guided image manipulation. 
Similarly, sampling in LCMs at $\tau_{i}$ starts with $\varepsilon \sim \mathcal{N}(\bm{0}, \bm{I})$ and updates:

\begin{equation}
\begin{aligned}
\label{eq:lcm-samp}
\hat{z}_{\tau_i} &= \sqrt{{\alpha}_{\tau_i}}z_0^{(\tau_{i+1})} + \sigma_{\tau_i} \varepsilon, \\
z_0^{(\tau_{i})} &= f_{\theta}(\hat{z}_{\tau_i}, \tau_i, c)
\end{aligned}
\end{equation}

\subsection{Inversion-Based Image Editing with LDMs}

DDIM inversion~\cite{song2020ddim} is effective for unconditional diffusion applications, but lacks consistency with additional text or image conditions.
As illustrated in Figure~\ref{fig::ddcm}a, the predicted $\bar{z}_0'$ deviates from the original source $z_0$, cumulatively leading to undesirable semantic changes.
This substantially restricts its use in image editing driven by natural language-guided diffusion.

\begin{figure*}
    \centering
    \begin{subfigure}[t]{.49\textwidth}
        \centering
        \includegraphics[width=1.02\linewidth]{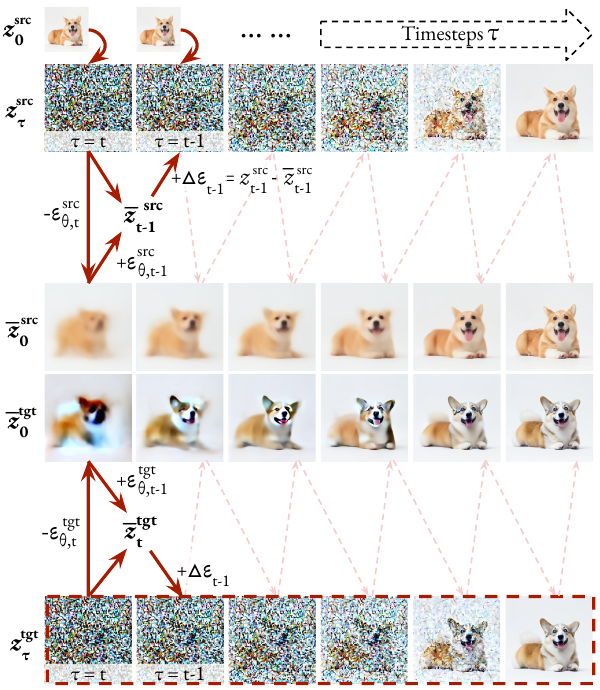}
        \caption{Dual-Branch inversion (Direct Inversion / CycleDiffusion) Editing.}
        \label{fig:cyclediff-editing}
    \end{subfigure}  
    ~
    \begin{subfigure}[t]{.49\textwidth}
        \centering
        \includegraphics[width=1.01\linewidth]{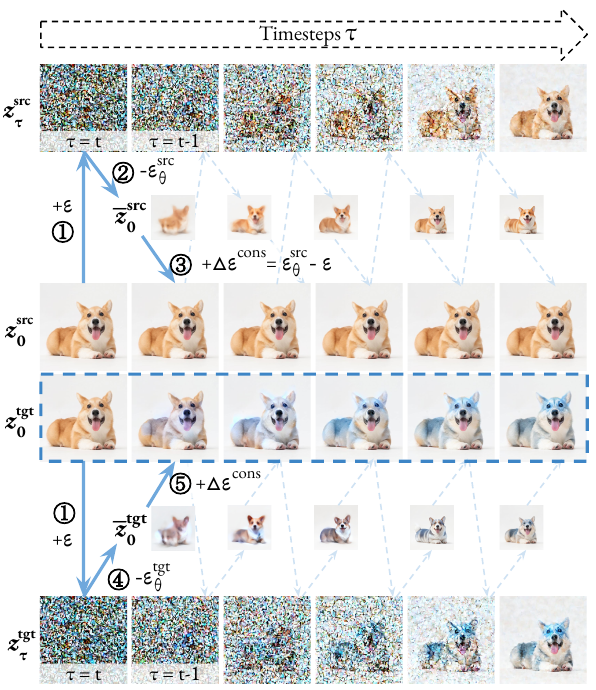}
        \caption{Inversion-free Editing with DDCM (Ours).}
        \label{fig::ddcm-editing}
    \end{subfigure}
    \vspace*{-5pt}
    \caption{A comparative overview of the dual-branch inversion editing and inversion-free editing enabled by DDCM. While the former iteratively calibrates $z_\tau^\textrm{tgt}$ in the target branch, inversion-free editing iteratively polishes the target branch initial $z_0^\textrm{tgt}$. In (b), we initialize $z_0^\textrm{tgt}$ with $z_0^\textrm{src}$ for visualization purposes, while in principle it can start from any random noise. The circled numbers correspond to Algorithm~\ref{alg:DDCM_Editing}. \vspace*{-15pt}}
    \label{fig:comparison}
\end{figure*}

To address this concern, various forms of inversion-based editing methods have been proposed.
The predominating approaches utilize \textit{optimization-based inversion}~\cite{mokady2023nti,li2023stylediffusion,dong2023prompt}.
These methods aim to ``correct'' the forward latents guided by the source prompt (referred to as the source branch) by aligning them with the DDIM inversion trajectory.
To tackle the efficiency bottlenecks and suboptimal consistency, very recent work has explored \textit{dual-branch inversion}~\cite{wu2023cyclediffusion,ju2023direct}.
These methods separate the source and target branches in the editing process: directly revert the source branch back to $z_0$ and iteratively calibrate the trajectory of the target branch.
As shown in Figure~\ref{fig:cyclediff-editing}, they calculate the distance between the source branch and the inversion branch (or directly sampled from $q$-sampling in~\cite{wu2023cyclediffusion}), and calibrate the target branch with this computed distance at each $t$.



\section{Denoising Diffusion Consistent Models}
\label{sec:DDCM}

We start with the following proposition.
\vspace*{-2pt}
\begin{prop}[Denoising Diffusion Consistent Models]
\label{prop:const}
Consider a special case of Eq (\ref{eq:fwd}) when $\sigma_t$ is chosen as $\sqrt{1 - \alpha_{t-1}}$ across all time $t$, the forward process naturally aligns with the Multistep (Latent) Consistency Sampling.
\end{prop}

\noindent
When $\sigma_t=\sqrt{1 - \alpha_{t-1}}$, the second term of Eq (\ref{eq:fwd}) vanishes:
\vspace*{-5pt}
\begin{equation}
\label{eq:ddcm1}
\begin{aligned}
    z_{t-1} = 
    & \sqrt{{{\alpha}}_{t-1}} \left( \frac{z_t - \sqrt{1 - {{\alpha}}_t} \varepsilon_{\theta}(z_t, t)}{\sqrt{{\alpha}}_t} \right) && \text{(predicted $z_0$)} \\
    & + \sqrt{1 - \alpha_{t-1}} \varepsilon_t\quad \varepsilon_t \sim \mathcal{N}(\bm{0},\bm{I}) && \text{(random noise)}
\end{aligned}
\end{equation}

\noindent
Consider $f(z_t, t; z_0) = \left( z_t - \sqrt{1 - {\alpha}_t} \varepsilon'(z_t,t;z_0) \right) / \sqrt{{\alpha}_t}$, where the initial $z_0$ is available (which is the case for image editing applications) and we replace the parameterized noise predictor  $\varepsilon_\theta$ with $\varepsilon'$ more generally.
Eq (\ref{eq:ddcm1}) becomes
\vspace*{-5pt}
\begin{equation}
\begin{aligned}
    z_{t-1} = \sqrt{{{\alpha}}_{t-1}} f(z_t,t;z_0) + \sqrt{1 - \alpha_{t-1}} \varepsilon_t
\end{aligned}
\end{equation}
which is in the same form as the Multistep Latent Consistency Sampling step in Eq (\ref{eq:lcm-samp}).

In order to make $f(z_t, t)$ self-consistent so that it can be considered as a consistency function, i.e., $f(z_t, t; z_0) = z_0$, we can directly solve the equation and $\varepsilon'$ can be computed without parameterization:
\vspace*{-5pt}
\begin{equation}
\label{eq:ddcm-eps}
\varepsilon^{\text{cons}} = \varepsilon'(z_t, t;z_0) = \frac{z_t - \sqrt{{\alpha}_t} z_0}{\sqrt{1 - {\alpha}_t}}
\end{equation}

As illustrated in Figure~\ref{fig::ddcm}c, we arrive at a non-Markovian forward process, in which $z_t$ directly points to the ground truth $z_0$ without neural prediction, and $z_{t-1}$ does not depend on the previous step $z_t$ like a consistency model.
We name this \textit{Denoising Diffusion Consistent Model} (DDCM).

\subsection{DDCM for Virtual Inversion}

We note that DDCM suggests an image reconstruction model without any explicit inversion operation, diverging from conventional DDIM inversion and its optimized or calibrated variations for image editing. 
It achieves the best efficiency as it allows the forward process to start from any random noise and supports multi-step consistency sampling. 
On the other hand, it ensures exact consistency between original and reconstructed images as each step on the forward branch $z_{t-1}$ only depends on the ground truth $z_0$ rather than the previous step $z_t$. 
Due to its inversion-free nature, we name this method \textit{Virtual Inversion}.
As outlined in Algorithm~\ref{alg:DDCM_inversion}, $z = z_0$ is ensured throughout the process without parameterization.

\vspace*{-10pt}
\begin{algorithm}[H]
    \begin{minipage}{\linewidth}
        \caption{DDCM Sampling for Virtual Inversion}
        \label{alg:DDCM_inversion}
        \begin{algorithmic}[1]
            \State \textbf{Input:} 
            \Statex \hskip\algorithmicindent Sequence of timesteps $\tau_{1}>\tau_2>\cdots >\tau_{N-1}$
            \Statex \hskip\algorithmicindent Reference initial input $z_0$
            \State Sample a random terminal noise $z_{\tau_1} \sim \mathcal{N}(\bm{0},\bm{I})$
            \State $\varepsilon^{\text{cons}}_{\tau_1} = (z_{\tau_1} - \sqrt{{\alpha}_{\tau_1}}z_0)/\sqrt{1-{\alpha}_{\tau_1}}$
            \State $z =  \left( z_{\tau_1} - \sqrt{1 - {\alpha}_{\tau_1}} \varepsilon^{\text{cons}}_{\tau_1} \right) / \sqrt{{\alpha}_{\tau_1}}$
            \For{$n=2$ to $N-1$}
                \State Sample noise $\varepsilon \sim \mathcal{N}(\bm{0},\bm{I})$
                \State $z_{\tau_n} = \sqrt{{\alpha}_{\tau_n}} z + \sqrt{1-{\alpha}_{\tau_n}} \varepsilon$
                \State $\varepsilon^{\text{cons}}_{\tau_n} = (z_{\tau_n} - \sqrt{{\alpha}_{\tau_n}}z_0)/\sqrt{1-{\alpha}_{\tau_n}}$
                \State $z = \left( z_{\tau_n} - \sqrt{1 - {\alpha}_{\tau_n}} \varepsilon^{\text{cons}}_{\tau_n} \right) / \sqrt{{\alpha}_{\tau_n}}$
            \EndFor
            \State \textbf{Output:} $z$
        \end{algorithmic} 
    \end{minipage}
\end{algorithm}

\subsection{DDCM for Inversion-Free Image Editing}

Existing inversion-based editing methods are limited for real-time and real-world language-driven image editing applications. 
First, most of them still depend on a time-consuming inversion process to obtain the inversion branch as a set of anchors.
Second, consistency remains a bottleneck given the efforts from optimization and calibration.
Recall that dual-branch inversion methods perform editing on the target branch by iteratively calibrating the ${z}_t^\textrm{tgt}$ with the actual distance between the source branch and the inversion branch at $t$, as is boxed in Figure~\ref{fig:cyclediff-editing}.
While they ensure faithful reconstruction by leaving the source branch untouched from the target branch, the calibrated $z_t^\textrm{tgt}$ does guarantee consistency from $z_t^\textrm{src}$ in the source branch, as can be seen from the visible difference between ${z}_0^\textrm{src}$ and ${z}_0^\textrm{tgt}$ in Figure~\ref{fig:cyclediff-editing}.
Third, all current inversion-based methods rely on variations of diffusion sampling, which are incompatible with efficient Consistency Sampling using LCMs.

DDCM offers an alternative to address these limitations, introducing an Inversion-Free Image Editing (InfEdit) framework.
While also adopting a dual-branch paradigm, the key of our InfEdit method is to directly calibrate the initial ${z}_0^\textrm{tgt}$ rather than the ${z}_t^\textrm{tgt}$ along the branch, as is boxed in Figure~\ref{fig::ddcm-editing}.
InfEdit starts from a random terminal noise $z^{\textrm{src}}_{\tau_1} = z^{\textrm{tgt}}_{\tau_1} \sim \mathcal{N}(\bm{0},\bm{I})$.
As shown in Figure~\ref{fig::ddcm-editing}, the source branch follows the DDCM sampling process without explicit inversion, and we directly compute the distance $\Delta\varepsilon^\textrm{cons}$ between $\varepsilon^\textrm{cons}$ the $\varepsilon_\theta^\textrm{src}$ (the predicted noise to reconstruct a $\bar{z}_0^\textrm{src}$).
For the target branch, we first compute the $\varepsilon_\theta^\textrm{tgt}$ to predict $\bar{z}_0^\textrm{tgt}$, and then calibrate the predicted target initial with the same $\Delta\varepsilon^\textrm{cons}$.
Algorithm~\ref{alg:DDCM_Editing} outlines the mathematical details of this process, in which we slightly abuse the notation to define $f_\theta(z_t,t,\varepsilon) = \left( z_t - \sqrt{1 - {\alpha}_t} \varepsilon \right) / \sqrt{{\alpha}_t}$.

\vspace*{-5pt}
\begin{algorithm}[H]
    \begin{minipage}{\linewidth}
        \caption{DDCM for inversion-free image editing}
        \label{alg:DDCM_Editing}
        \begin{algorithmic}[1]
            \Statex \textbf{Input:} 
            \Statex \hskip\algorithmicindent Conditional Diffusion/Consistency Model $\varepsilon_\theta(\cdot, \cdot, \cdot)$
            \Statex \hskip\algorithmicindent Sequence of timesteps $\tau_{1}>\tau_2>\cdots >\tau_{N-1}$
            \Statex \hskip\algorithmicindent Reference initial input $z_0^{\textrm{src}}$
            \Statex \hskip\algorithmicindent Source/target prompts as conditions $c^{\text{\textrm{src}}}, c^\text{\textrm{tgt}}$
            \State Sample a random terminal noise $z^{\textrm{src}}_{\tau_1} = z^{\textrm{tgt}}_{\tau_1} \sim \mathcal{N}(\bm{0},\bm{I})$
            \State $\varepsilon^{\text{cons}}_{\tau_1} = (z^{\textrm{src}}_{\tau_1} - \sqrt{{\alpha}_{\tau_1}}z_0^{\textrm{src}})/\sqrt{1-{\alpha}_{\tau_1}}$
            \State $\varepsilon^{\textrm{src}}_{\tau_1}, \varepsilon^{\textrm{tgt}}_{\tau_1} = \varepsilon_\theta(z^{\textrm{src}}_{\tau_1}, {\tau_1}, c^{\textrm{src}}), \varepsilon_\theta(z^{\textrm{tgt}}_{\tau_1}, {\tau_1}, c^{\textrm{tgt}})$
            \State $z_0^{\textrm{tgt}} = f_\theta(z^{\textrm{tgt}}_{\tau_1}, {\tau_1}, \varepsilon^{\textrm{tgt}}_{\tau_1} - \varepsilon^{\textrm{src}}_{\tau_1} + \varepsilon^{\text{cons}}_{\tau_1})$
            \For{$n=2$ to $N-1$}
                \State Sample noise \textcolor{red}{$\varepsilon \sim \mathcal{N}(\bm{0},\bm{I})$}
                \State {\textcircled{1}} $z^{\textrm{src}}_{\tau_n} = \sqrt{{\alpha}_{\tau_n}} z_0^{\textrm{src}} + $\textcolor{red}{$\sqrt{1-{\alpha}_{\tau_n}} \varepsilon$}
                \State {\textcircled{1}} $z^{\textrm{tgt}}_{\tau_n} = \sqrt{{\alpha}_{\tau_n}} z_0^{\textrm{tgt}} + $\textcolor{red}{$\sqrt{1-{\alpha}_{\tau_n}} \varepsilon$}
                \State {\textcircled{2}} $\varepsilon^{\textrm{src}}_{\tau_n} = \varepsilon_\theta(z^{\textrm{src}}_{\tau_n}, {\tau_n}, c^{\textrm{src}})$
                \State {\textcircled{3}} $\varepsilon^{\text{cons}}_{\tau_n} = (z^{\textrm{src}}_{\tau_n} - \sqrt{{\alpha}_{\tau_n}}z_0^{\textrm{src}})/\sqrt{1-{\alpha}_{\tau_n}}$
                \State {\textcircled{4}}$^* \varepsilon^{\textrm{tgt}}_{\tau_n} = \varepsilon_\theta(z^{\textrm{tgt}}_{\tau_n}, {\tau_n}, c^{\textrm{tgt}})$
                \State {\textcircled{5}} $z_0^{\textrm{tgt}} = f_\theta(z^{\textrm{tgt}}_{\tau_n}, {\tau_n}, \varepsilon^{\textrm{tgt}}_{\tau_n} - \varepsilon^{\textrm{src}}_{\tau_n} + \varepsilon^{\text{cons}}_{\tau_n})$
            \EndFor
            \State \textbf{Output:} $z_0^{\textrm{tgt}}$
            \State \textcolor{gray}{*\textit{Vanilla target noise prediction, no attention control}}.
        \end{algorithmic} 
    \end{minipage}
\end{algorithm}

\vspace*{-5pt}
InfEdit addresses the current limitations of inversion-based editing methods.
First, DDCM sampling allows us to abandon the inversion branch anchors required by previous methods, saving a significant amount of computation.
Second, the current dual-branch methods calibrate $z_t^\textrm{tgt}$ over time, while InfEdit directly refines the predicted initial $z_0^\textrm{tgt}$, without suffering from the cumulative errors over the course of sampling.
Third, our framework is compatible with efficient Consistency Sampling using LCMs, enabling efficient sampling of the target image within very few steps.

\section{Unifying Attention Control for Language-Guided Editing}
\label{sec:UAC}

InfEdit suggests a general inversion-free framework for image editing motivated by DDCM.
In the realm of language-driven editing, achieving a nuanced understanding of the language condition and facilitating finer-grained interaction across modalities becomes a challenge.
\citet{hertz2023p2p} noticed that the interaction between the text and image modalities occurs in the parameterized noise prediction network $\varepsilon_\theta$, and opened up a series of attention control methods to compute a noise $\widehat{\varepsilon_\theta^\textrm{tgt}}$ that more accurately aligns with the language prompts.
In the context of InfEdit specifically, attention control refines the original predicted target noise $\varepsilon_\theta^\textrm{tgt}$ (noted in \textcircled{4} in Algorithm~\ref{alg:DDCM_Editing} and Figure~\ref{fig::ddcm-editing}) with $\widehat{\varepsilon_\theta^\textrm{tgt}}$.

We follow~\cite{hertz2023p2p} in terms of notation.
Each basic block of the U-Net noise predictor contains a cross-attention module and a self-attention module.
The spatial features are linearly projected into queries ($Q$).
In cross-attention, the text features are linearly projected into keys ($K$) and values ($V$).
In self-attention, the keys ($K$) and values ($V$) are also obtained from linearly projected spatial features.
The attention mechanism~\cite{vaswani2017attention} can be given as:

\vspace*{-15pt}
\begin{equation}
\text{Attention}(K,Q,V) = \textcolor{red}{M}V = \textcolor{red}{\textrm{softmax}\left(\frac{QK^T}{\sqrt{d}}\right)}V
\vspace*{-3pt}
\end{equation}

\noindent
in which \(M_{i,j}\) represents the attention map that determines the weight to aggregate the value of the \(j\)-th token on pixel \(i\), and \(d\) denotes the dimension for $K$ and $Q$.

Natural language specifies a wide spectrum of semantic changes.
In the following, we describe how \textit{rigid semantic changes}, e.g., those on the visual features and background, can be controlled via cross attention~\cite{hertz2023p2p}; and how \textit{non-rigid semantic changes}, e.g., those leading to adding/removing an object, novel action manners and physical state changes of objects, can be controlled via mutual self-attention~\cite{cao2023masactrl}.
We then introduce a Unified Attention Control (UAC) protocol for both types of semantic changes.

\subsection{Cross-Attention Control}

Prompt-to-Prompt (P2P)~\cite{hertz2023p2p} observed that cross-attention layers can capture the interaction between the spatial structures of pixels and words in the prompts, even in early steps.
This finding makes it possible to control the cross-attention for editing rigid semantic changes, simply by replacing the cross-attention map of generated images with that of the original images.

\vspace*{-5pt}
\paragraph{Global Attention Refinement}
At time step $t$, we compute the attention map \(M_{t}\) averaged over layers given the noised latent $z_t$ and the prompt for both source and target branch.
We drop the time step for simplicity and represent the source and target attention maps as $M^{\textrm{src}}$ and $M^{\textrm{tgt}}$.
To represent the common details, an alignment function $A(i) = j$ is introduced which signifies that the $i^{\textrm{th}}$ word in the target prompt corresponds to the  $j^{\textrm{th}}$ word in the source prompt.
Following \citet{hertz2023p2p}, we refine the target attention map by injecting the source attention map over the common tokens.

\vspace*{-15pt}
\begin{equation}
    \textrm{Refine}(M^{\textrm{src}},M^{\textrm{tgt}})_{i,j}
    = \begin{cases}
        \left(M^{\textrm{tgt}}\right)_{i,j} 
        & \text{if} A(j)=\text{None} \\ 
        \left(M^{\textrm{src}}\right)_{i,A(j)} 
        & \text{otherwise}
    \end{cases}
\end{equation}
\vspace*{-10pt}

This ensures that the common information from the source prompt is accurately transferred to the target, while the requested changes are made.

\vspace*{-5pt}
\paragraph{Local Attention Blends}
Besides global attention refinement, we adapt the blended diffusion mechanism from \citep{avrahami2022blended,hertz2023p2p}.
Specifically, the algorithm takes optional inputs of target blend words $w^{\textrm{tgt}}$, which are words in the target prompt whose semantics need to be added; and source blend words $w^{\textrm{src}}$, which are words in the source prompt whose semantics need to be preserved. 
At time step $t$, we blend the noised target latent $z_t^{\textrm{tgt}}$ following:

\vspace*{-15pt}
\begin{equation}
\begin{aligned}
    m^{\textrm{tgt}} &= \text{Threshold} \big[M_t^{\textrm{tgt}}(w^{\textrm{tgt}}), a^{\textrm{tgt}}\big] \\
    m^{\textrm{src}} &= \text{Threshold} \big[M_t^{\textrm{src}}(w^{\textrm{src}}), a^{\textrm{src}}\big] \\
    z_t^{\textrm{tgt}} &= (1-m^{\textrm{tgt}}+m^{\textrm{src}}) \odot z_t^{\textrm{src}} + (m^{\textrm{tgt}}-m^{\textrm{src}}) \odot z_t^{\textrm{tgt}}  \\
\end{aligned}
\end{equation}
\vspace*{-10pt}

\noindent
in which $m^{\textrm{tgt}}$ and $m^{\textrm{src}}$ are binary masks obtained by calibrating the aggregated attention maps $M_t^{\textrm{tgt}}(w^{\textrm{tgt}}),M_t^{\textrm{src}}(w^{\textrm{src}}) $ with threshold parameters $a^{\textrm{tgt}}$ and $a^{\textrm{src}}$ using threshold function:

\vspace*{-5pt}
\begin{equation}
\text{Threshold}(M,a)_{i,j} = 
\begin{cases}
    1\quad  M_{i,j}\geq a  \\
    0\quad  M_{i,j}<a 
\end{cases}
\end{equation}

\vspace*{-15pt}
\paragraph{Scheduling Cross-Attention Control}
Applying cross-attention control throughout the entire sampling schedule will overly focus on spatial consistency, leading to an inability to capture the intended changes.
Follow~\cite{hertz2023p2p}, we perform cross-attention control only in early steps before $\tau_c$, interpreted as the cross-attention control strength:

\vspace*{-15pt}
\begin{equation*}
    \textrm{CrossEdit}(M^{\textrm{src}},M^{\textrm{tgt}},t):=\begin{cases}\textrm{Refine}(M^{\textrm{src}},M^{\textrm{tgt}}) &t \geq \tau_c \\ M^{\textrm{tgt}} & t < \tau_c\end{cases}
\vspace*{-5pt}
\end{equation*}

\subsection{Mutual Self-Attention Control}

One key limitation of cross-attention control lies in its inability in non-rigid editing. 
Instead of applying controls over the cross-attention modules, MasaCtrl~\cite{cao2023masactrl} observed that the layout of the objects can be roughly formed in the self-attention queries, covering the non-rigid semantic changes complying with the target prompt. 
The core idea is to synthesize the structural layout with the target prompt in the early steps with the original $Q^{\textrm{tgt}}, K^{\textrm{tgt}}, V^{\textrm{tgt}}$ in the self-attention; and then to query semantically similar contents in $K^{\textrm{src}}, V^{\textrm{src}}$ with the target query $Q^{\textrm{tgt}}$.

\vspace*{-5pt}
\paragraph{Controlling Non-Rigid Semantic Changes}
MasaCtrl suffers from the issue of undesirable non-rigid changes.
As shown in Figure~\ref{fig::qualitative}, MasaCtrl can lead to significant inconsistency from the source images, especially in terms of the composition of objects and when there are multiple objects and complex backgrounds.
This is not surprising, as the target query $Q^{\textrm{tgt}}$ is used throughout the self-attention control schedule.
Instead of relying on the target prompts to guide the premature steps, we form the structural layout with the source self-attention $Q^{\textrm{src}}, K^{\textrm{src}}, V^{\textrm{src}}$ in the self-attention.
We show in Section~\ref{sec:experiments} that this design enables high-quality non-rigid changes while maintaining satisfying structural consistency.

\vspace*{-5pt}
\paragraph{Scheduling Mutual Self-Attention Control}
This mutual self-attention control is applied in the later steps after $\tau_s$, interpreted as the mutual self-attention control strength:

\vspace*{-15pt}
\begin{equation}
    \begin{aligned}
        \textrm{SelfEdit}(\{Q^{\textrm{src}},K^{\textrm{src}},V^{\textrm{src}}\},\{Q^{\textrm{tgt}},K^{\textrm{tgt}},V^{\textrm{tgt}}\},t) := \\
        \begin{cases}\{Q^{\textrm{src}},K^{\textrm{src}},V^{\textrm{src}}\}&t \geq \tau_s \\ \{Q^{\textrm{tgt}},K^{\textrm{src}},V^{\textrm{src}}\} & t < \tau_s\end{cases}\\
    \end{aligned}
\end{equation}

\subsection{Unified Attention Control}

\begin{figure}[!t]
    \centering
    \includegraphics[width=1.0\linewidth]{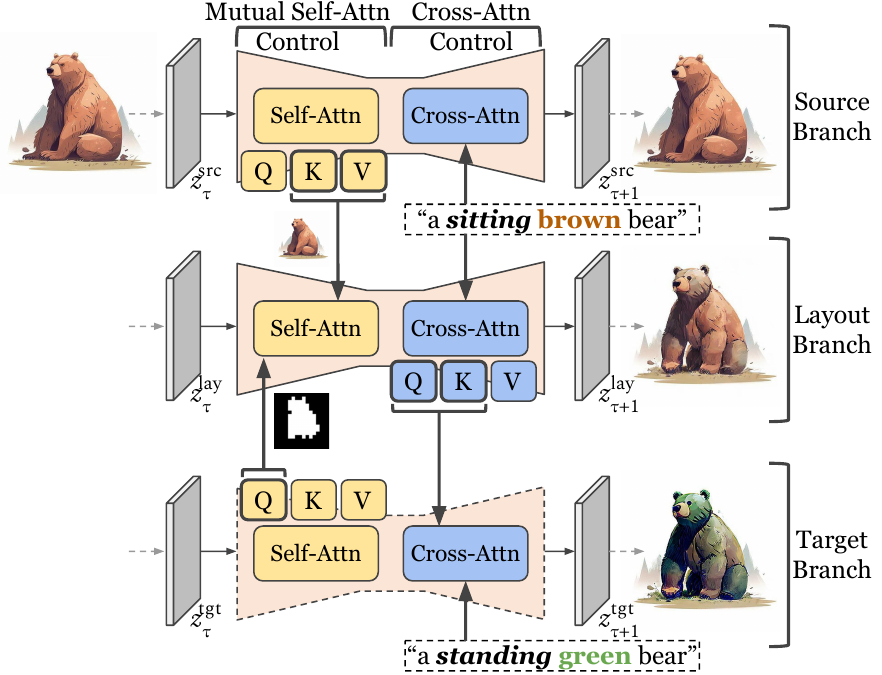}
    \vspace*{-15pt}
    \caption{The proposed United Attention Control (UAC) framework to unify cross-attention control and mutual self-attention control. UAC introduces an additional layout branch as an intermediate to host the desired composition and structural information in the target image. \vspace*{-15pt}}
    \label{fig::uac}    
\end{figure}

To enable both rigid and non-rigid semantic changes within one unified framework is not trivial.

\begin{figure}[!h]
    \centering
    \vspace*{-10pt}
    \begin{subfigure}[t]{.105\textwidth}
        \centering
        \includegraphics[width=1.00\linewidth]{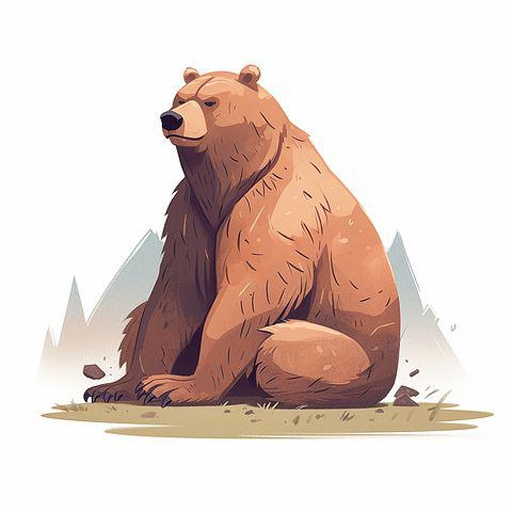}
        \vspace*{-15pt}
        \caption{The input source image.}
        \label{fig:src}
    \end{subfigure}
    ~    
    \begin{subfigure}[t]{.105\textwidth}
        \centering
        \includegraphics[width=1.00\linewidth]{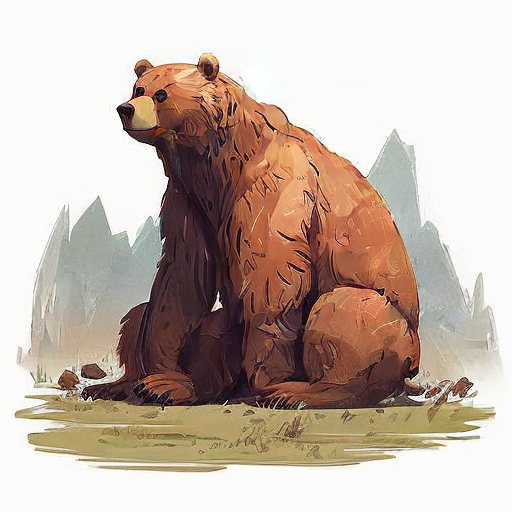}
        \vspace*{-15pt}
        \caption{2-branch target output.}
        \label{fig:2-tgt}
    \end{subfigure}
    ~    
    \begin{subfigure}[t]{.105\textwidth}
        \centering
        \includegraphics[width=1.00\linewidth]{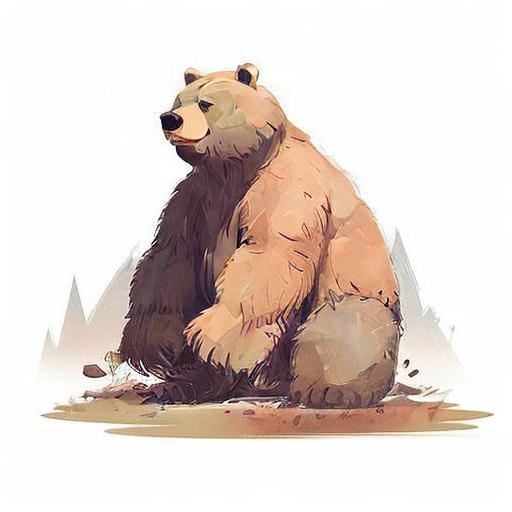}
        \vspace*{-15pt}
        \caption{3-branch layout ouput.}
        \label{fig:lay}
    \end{subfigure}
    ~    
    \begin{subfigure}[t]{.105\textwidth}
        \centering
        \includegraphics[width=1.00\linewidth]{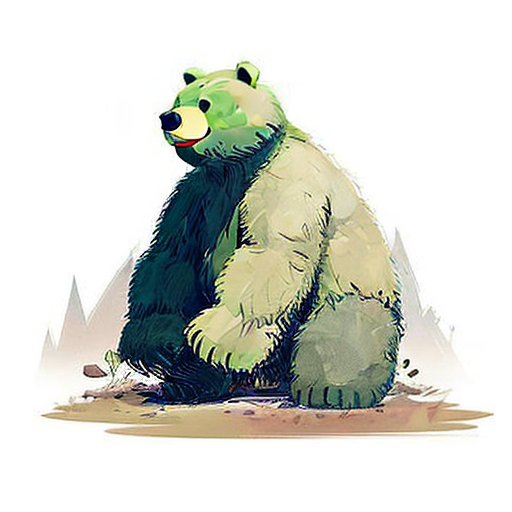}
        \vspace*{-15pt}
        \caption{3-branch target output.}
        \label{fig:tgt}
    \end{subfigure}
    \vspace*{-5pt}
    \caption{A comparison of the target branch outputs to edit ``a \textbf{sitting} \textbf{brown} bear'' to ``a \textbf{standing} \textbf{green} bear'', involving both rigid and non-rigid semantic transformations. Random seed is fixed. \vspace*{-10pt}}
    \label{fig:branch}
\end{figure}

As is illustrated in Figure~\ref{fig:branch}, the na\"ive combination of cross-attention control and mutual self-attention control sequentially would lead to a sub-optimal outcome in the original dual-branch setup, especially failing the global attention refinement.
To address this issue, we introduce the Unified Attention Control (UAC) framework.
UAC unifies cross-attention control and mutual self-attention control with an additional latent \textit{layout branch}, which serves as an intermediate to host the desired composition and structural information in the target image.

The UAC framework is detailed in Algorithm~\ref{alg:UAC} and illustrated in Figure~\ref{fig::uac}.
During each forward step of the diffusion process, UAC starts with mutual self-attention control on $z^{\textrm{src}}$ and $z^{\textrm{tgt}}$ and assigns the output to the layout branch latent $z^{\textrm{lay}}$. 
Following this, cross-attention control is applied on $M^{\textrm{lay}}$ and $M^{\textrm{tgt}}$ to refine the semantic information for $M^{\textrm{tgt}}$. 
As is shown in Figure~\ref{fig:lay}, the layout branch output $z_0^{\textrm{lay}}$ reflects the requested non-rigid changes (e.g., ``standing''), while preserving the non-rigid content semantics (e.g., ``brown'').
The target branch output $z_0^{\textrm{tgt}}$ (Figure~\ref{fig:tgt}) builds upon the structural layout of the $z_0^{\textrm{lay}}$ while reflecting the requested non-rigid changes (e.g., ``green'').

\begin{algorithm}[h]
    \begin{minipage}{\linewidth}
        \caption{Unified Attention Control Image Editing}\label{alg:UAC}
        \begin{algorithmic}[1]
            \State \textbf{Input:}
            \Statex \hskip\algorithmicindent Conditional Diffusion/Consistency Model $\varepsilon_\theta(\cdot, \cdot, \cdot)$
            \Statex \hskip\algorithmicindent Current timestep $\tau$
            \Statex \hskip\algorithmicindent Reference initial input $z_0^{\textrm{src}}$
            \Statex \hskip\algorithmicindent Source/target prompts as conditions $c^{\text{\textrm{src}}}, c^\text{\textrm{tgt}}$
            \Statex \hskip\algorithmicindent Source/target blend words $w^{\text{\textrm{src}}}, w^\text{\textrm{tgt}}$
            \Statex \hskip\algorithmicindent Input latents $z^{\textrm{src}}_\tau, z^{\textrm{tgt}}_\tau, z^{\textrm{lay}}_\tau $
            \State $\varepsilon^{\textrm{src}},\{Q^{\textrm{src}},K^{\textrm{src}},V^{\textrm{src}}\}, M^{\textrm{src}} = \varepsilon_\theta(z_{\tau}^{\textrm{src}},\tau,c^{\textrm{src}})$
            \State $\varepsilon^{\textrm{tgt}},\{Q^{\textrm{tgt}},K^{\textrm{tgt}},V^{\textrm{tgt}}\}, M^{\textrm{tgt}} = \varepsilon_\theta(z_{\tau}^{\textrm{tgt}},\tau,c^{\textrm{tgt}})$
            \State $\{\widehat{Q^{\textrm{lay}}},\widehat{K^{\textrm{lay}}},\widehat{V^{\textrm{lay}}}\} = $
            \State \quad \quad$ \textrm{SelfEdit}(\{Q^{\textrm{src}},K^{\textrm{src}},V^{\textrm{src}}\},\{Q^{\textrm{tgt}},K^{\textrm{tgt}},V^{\textrm{tgt}}\},\tau)$
            \State $\varepsilon^{\textrm{lay}}, M^{\textrm{lay}} = \varepsilon_\theta(z_{\tau}^{\textrm{lay}},\tau,c^{\textrm{src}};\{\widehat{Q^{\textrm{lay}}},\widehat{K^{\textrm{lay}}},\widehat{V^{\textrm{lay}}}\})$
            \State $\widehat{M^{\textrm{tgt}}} = \textrm{CrossEdit}(M^{\textrm{lay}},M^{\textrm{tgt}},\tau)$
            \State $\widehat{\varepsilon^{\textrm{tgt}}}= \varepsilon_\theta(z_{\tau}^{\textrm{tgt}},\tau,c^{\textrm{tgt}}; \widehat{M^{\textrm{tgt}}}) $
            \State $z_{\tau+1}^{\textrm{src}},z_{\tau+1}^{\textrm{tgt}},z_{\tau+1}^{\textrm{lay}} = $
            \State \quad \quad $\textrm{Sample}([z_{\tau}^{\textrm{src}},z_{\tau}^{\textrm{tgt}},z_{\tau}^{\textrm{lay}}],[\varepsilon^{\textrm{src}},\widehat{\varepsilon^{\textrm{tgt}}},\varepsilon^{\textrm{lay}}],\tau) $
            \State $m^{\textrm{tgt}} = \text{Threshold} \big[M_\tau^{\textrm{tgt}}(w^{\textrm{tgt}}), a^{\textrm{tgt}}\big]$ 
            \State $m^{\textrm{src}} = \text{Threshold} \big[M_\tau^{\textrm{src}}(w^{\textrm{src}}), a^{\textrm{src}}\big]$ 
            \State $z_{\tau+1}^{\textrm{tgt}} = (1-m^{\textrm{tgt}}+m^{\textrm{src}}) \odot z_{\tau+1}^{\textrm{src}} + (m^{\textrm{tgt}}-m^{\textrm{src}}) \odot z_{\tau+1}^{\textrm{tgt}}$
            \State \textbf{Output:} $z_{\tau+1}^{\textrm{src}},z_{\tau+1}^{\textrm{tgt}},z_{\tau+1}^{\textrm{lay}}$
        \end{algorithmic} 
    \end{minipage}
\end{algorithm}
\vspace*{-10pt}

\section{Experiments}
\label{sec:experiments}

\subsection{Experiment Setups}

\begin{figure*}[!htp]
    \centering
    \includegraphics[width=1.0\linewidth]{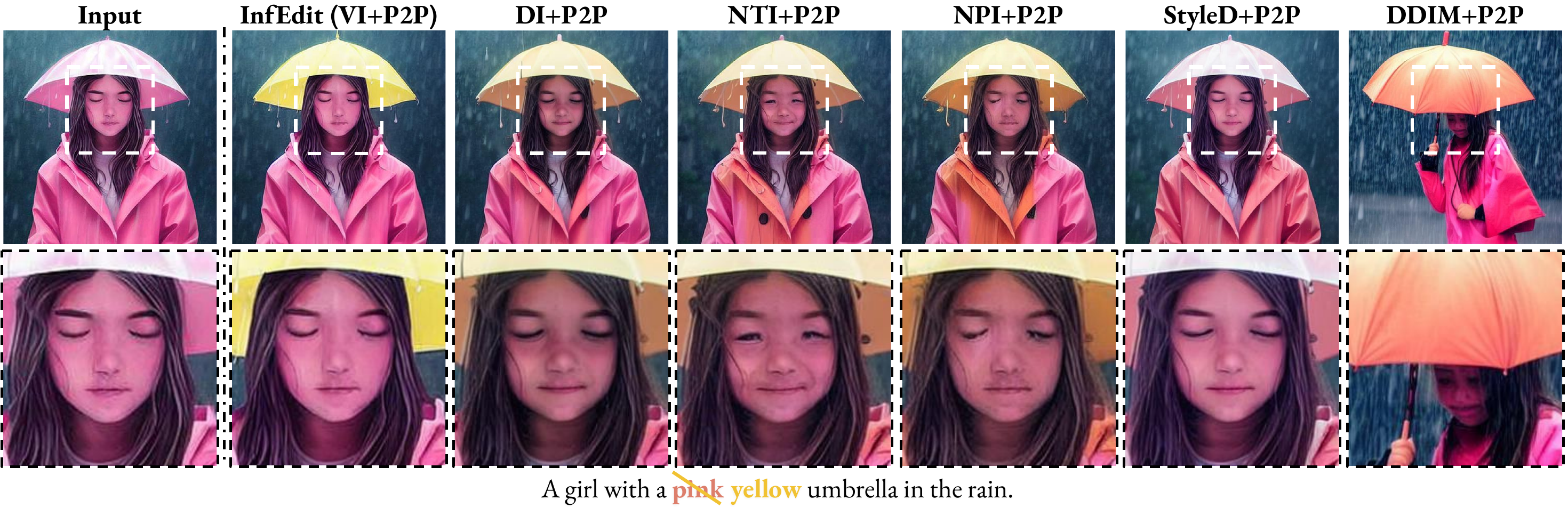}
    \vspace*{-20pt}
    \caption{A qualitative example for ablation over inversion methods. With the same P2P attention control, InfEdit (VI+P2P) allows faithful semantic changes as well as better consistency. \vspace*{-5pt}}
    \label{fig::comp-inv}
\end{figure*}
\begin{table*}[htbp]
\small
\centering
  \renewcommand\arraystretch{1.0}
    \setlength{\tabcolsep}{0.7mm}{
    \begin{threeparttable}
    \begin{tabular}{cccrrrcccrrc}
    \toprule
    \multicolumn{2}{c}{\textbf{Method}} & \multicolumn{1}{c}{\textbf{Structure}} & \multicolumn{4}{c}{\textbf{Background Preservation}} & \multicolumn{2}{c}{\textbf{CLIP Similarity}} & \multicolumn{3}{c}{\textbf{Efficiency (sec / \#)}} \\ 
    \cmidrule(r){1-2} \cmidrule(r){3-3} \cmidrule(r){4-7} \cmidrule(r){8-9} \cmidrule(r){10-12} 
    \textbf{Inverse} & \textbf{Edit} & \textbf{Distance}$_{10^3}$ $\downarrow$ & \textbf{PSNR} $\uparrow$ & \textbf{LPIPS}$_{10^3}^\downarrow$ & \textbf{MSE}$_{10^4}^\downarrow$ & \textbf{SSIM}$_{10^2}^\uparrow$ & \textbf{Whole} $\uparrow$ & \textbf{Edited} $\uparrow$ & \textbf{Inverse Time}$\downarrow$ & \textbf{Forward Time}$\downarrow$ & \textbf{Steps}$\downarrow$\\
    \cmidrule(r){1-2} \cmidrule(r){3-3} \cmidrule(r){4-7} \cmidrule(r){8-9} \cmidrule(r){10-12} 
    \textbf{DDIM}& \textbf{P2P} & 69.43 & 17.87 & 208.80 & 219.88 & 71.14 & 25.01 & 22.44 & 10.93 $\pm$ 0.01 & 12.79 $\pm$ 0.01 & 50 \\
    \cmidrule(r){1-2} \cmidrule(r){3-3} \cmidrule(r){4-7} \cmidrule(r){8-9} \cmidrule(r){10-12} 
    \textcolor{gray}{\textbf{CycleD}} & \textcolor{gray}{\textbf{P2P}} & \textcolor{gray}{6.06} & \textcolor{gray}{28.25} & \textcolor{gray}{43.96} & \textcolor{gray}{25.85} & \textcolor{gray}{85.61} & \textcolor{red}{23.68} & \textcolor{red}{20.87} & \textcolor{gray}{N/A} & \textcolor{gray}{4.55 $\pm$ 0.02} & \textcolor{gray}{32} \\
    \textbf{NT}& \textbf{P2P} & 13.44 & 27.03 & 60.67 & 35.86 & 84.11 & 24.75 & 21.86 & 132.39 $\pm$ 7.69 & 12.90 $\pm$ 0.01 & 50 \\
    \textbf{NP}& \textbf{P2P} & 16.17 &  26.21 &  69.01 &  39.73 & 83.40 & 24.61 & 21.87 & 4.14 $\pm$ 0.00 & 12.78 $\pm$ 0.01 & 50 \\
    \textbf{StyleD}& \textbf{P2P} & \textbf{11.65} & 26.05 & 66.10 & 38.63 & 83.42 & 24.78 & 21.72 & 810.17 $\pm$ 7.77 & 28.18 $\pm$ 1.30 & 50 \\
    \textbf{DI} & \textbf{P2P} & \textbf{11.65} & 27.22 & 54.55 & \textbf{32.86}  & 84.76 & \textbf{25.02} & \textbf{22.10} & 16.83 $\pm$ 0.02 & 12.87 $\pm$ 0.01 & 50 \\
    \rowcolor[HTML]{C9DAF8}
    \textbf{VI} & \textbf{P2P} & 14.22 & \textbf{27.52} & \textbf{47.98} & 34.17 & \textbf{85.05} & 24.89 & 22.03 & \textbf{N/A} & \textbf{4.50} $\pm$ 0.01 & \textbf{32} \\
    \cmidrule(r){1-2} \cmidrule(r){3-3} \cmidrule(r){4-7} \cmidrule(r){8-9} \cmidrule(r){10-12} 
    \rowcolor[HTML]{C9DAF8}
    \textbf{VI*} & \textbf{P2P} & 15.61 & 26.64 & 55.85 & 41.15  & 84.66 & 24.57 & 21.69 & \textbf{N/A} & 2.60 $\pm$ 0.00 & 15 \\
    \rowcolor[HTML]{C9DAF8}
    \textbf{VI*} & \textbf{UAC} & 13.78 & \textbf{28.51} & \textbf{47.58} & \textbf{32.09} & \textbf{85.66} & \textbf{25.03} & \textbf{22.22} & \textbf{N/A} & \textbf{2.22} $\pm$ 0.02 & \textbf{12} \\
    \bottomrule
    \end{tabular}
    \begin{tablenotes}
        \item[*] Using the Latent Consistency Model (LCM) as the base model. Otherwise, Stable Diffusion (SD) v1.4 is adopted.
    \end{tablenotes}
    \end{threeparttable}}
    \vspace{-10pt}
    \caption{Aggregated performances of different image inversion and editing methods on PIE-bench. We break InfEdit into the Virtual inversion (VI) and arbitrary choices of attention control mechanism and diffusion backbone. VI competes and even surpasses other inversion methods with the same P2P attention control. The integration of unified attention control (UAC) and the LCM backbone further enhances its performance. Notably, InfEdit runs about an order of magnitude faster than most of the baselines on one single A40. \vspace{-10pt}}
    \label{tab:inversion_based_editing}
\end{table*}

\paragraph{Benchmarks}
We used established benchmarks to evaluate our proposed image editing method:
\begin{itemize}[leftmargin=*]
    \setlength\itemsep{-0.05em}
    \item \textbf{Language-Guided Image Editing.} We evaluate on the PIE-Bench introduced by~\citet{ju2023direct}, which assesses language-guided image editing in 9 different scenarios.
    \item \textbf{Image-to-Image (I2I) Translation.} We also evaluate on the I2I tasks at the scene-level (Summer$\leftrightarrow$Winter) and object-level (Horse$\leftrightarrow$Zebra)~\cite{zhu2017unpaired}.
\end{itemize}

\paragraph{Evaluation Metrics}
We employ 4 distinct evaluation metrics to assess the generated image's quality, the accuracy of the translation, the consistency against the source images, and the efficiency of the editing process.
\begin{itemize}[leftmargin=*]
    \setlength\itemsep{-0.1em}
    \item \textbf{Image Quality.} We use the Fréchet Inception Distance (FID)~\citep{heusel2017FID} score, which compares the model outputs to real image distributions;
    \item \textbf{Translation Quality.} We use the CLIPScore~\citep{hessel2021clipscore} to quantify the semantic similarity of the generated image and target prompt with CLIP~\citep{radford2021learning}.
    \item \textbf{Translation Consistency.} We measure translation consistency using four different metrics: Mean Squared Error (MSE), Peak Signal-to-Noise Ratio (PSNR), Structural Similarity Index Measure (SSIM)~\citep{wang2004image}, and Learned Perceptual Image Patch Similarity (LPIPS)~\cite{zhang2018unreasonable}.
    \item \textbf{Efficiency.} We directly compare the computation time on one A40 GPU for the inversion and forward process, as well as the number of sampling steps.
\end{itemize}

\subsection{Inversion v.s. Inversion-Free Comparison}
\label{subsec:inversion-editing}

\begin{figure*}
    \centering
    \begin{subfigure}[t]{.43\textwidth}
        \centering
        \includegraphics[width=1.\linewidth]{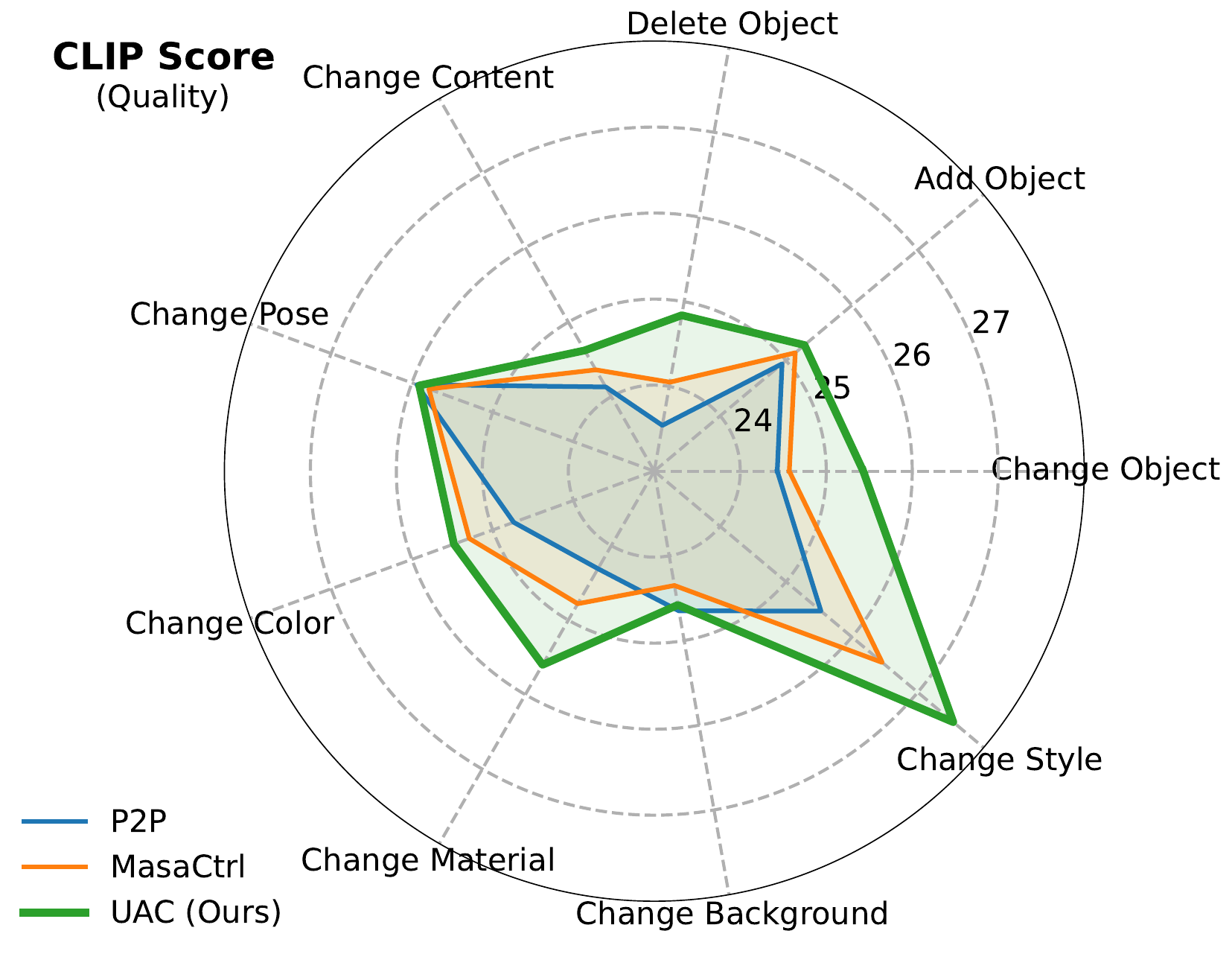}
        \caption{Spider chart for editing quality.}
        \label{fig:quality}
    \end{subfigure}
    ~ 
    \hspace*{-15pt}
    \begin{subfigure}[t]{.43\textwidth}
        \centering
        \includegraphics[width=1.\linewidth]{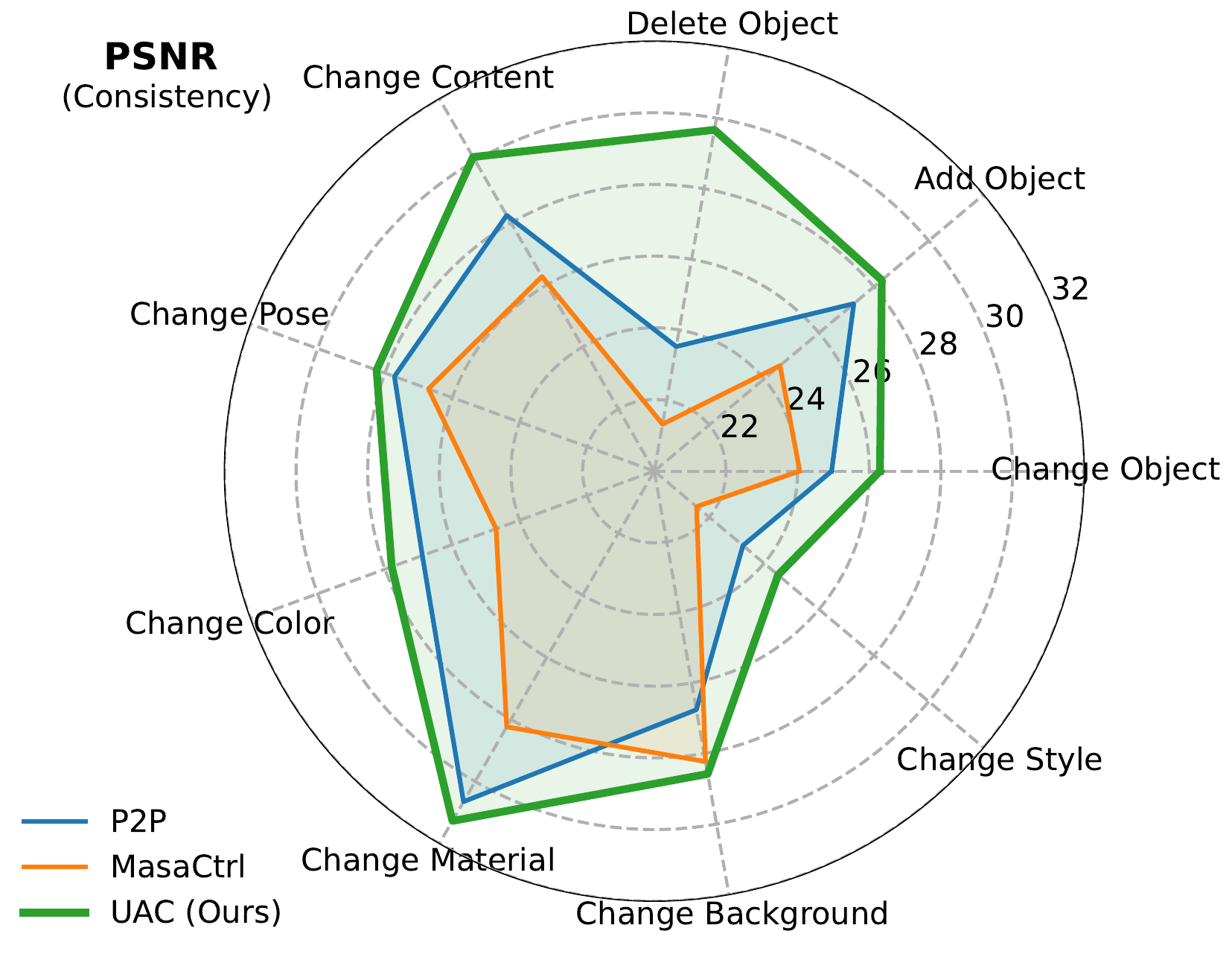}
        \caption{Spider chart for editing consistency.}
        \label{fig:consistency}
    \end{subfigure}   
    ~
    \hspace*{-10pt}
    \begin{subfigure}[t]{.11\textwidth}
        \centering
        \includegraphics[width=1.\linewidth]{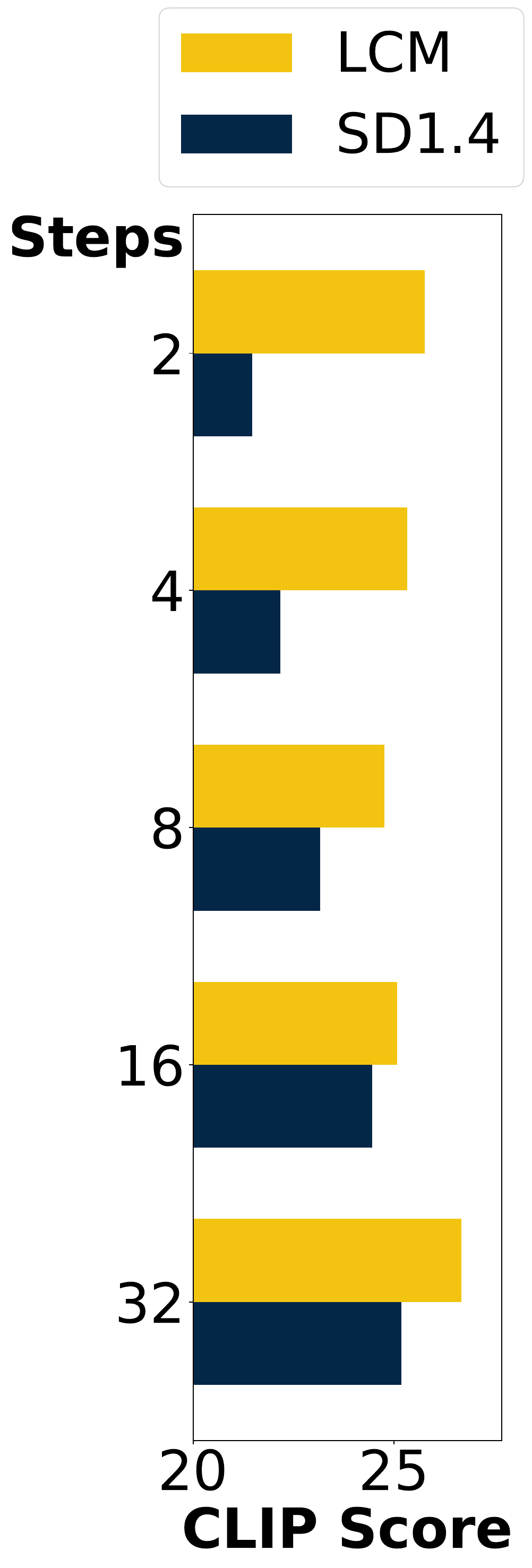}
        \caption{Efficiency.}
        \label{fig:efficiency}
    \end{subfigure}
    \vspace*{-10pt}
    \caption{A comprehensive performance evaluation on the PIE-bench. We present spider charts of editing quality (CLIP Scores) and consistency (PSNR) across 9 editing tasks for Prompt-to-Prompt (P2P), MasaCtrl, and Unified Attention Control (UAC) methods. Accompanied by an analysis of editing efficiency for Stable Diffusion (SD) 1.4 and Latent Consistency Model (LCM) across different steps. \vspace*{-10pt}}
    \label{fig:comparison}
\end{figure*}

\begin{figure*}[!htp]
    \centering
    \includegraphics[width=1.0\linewidth]{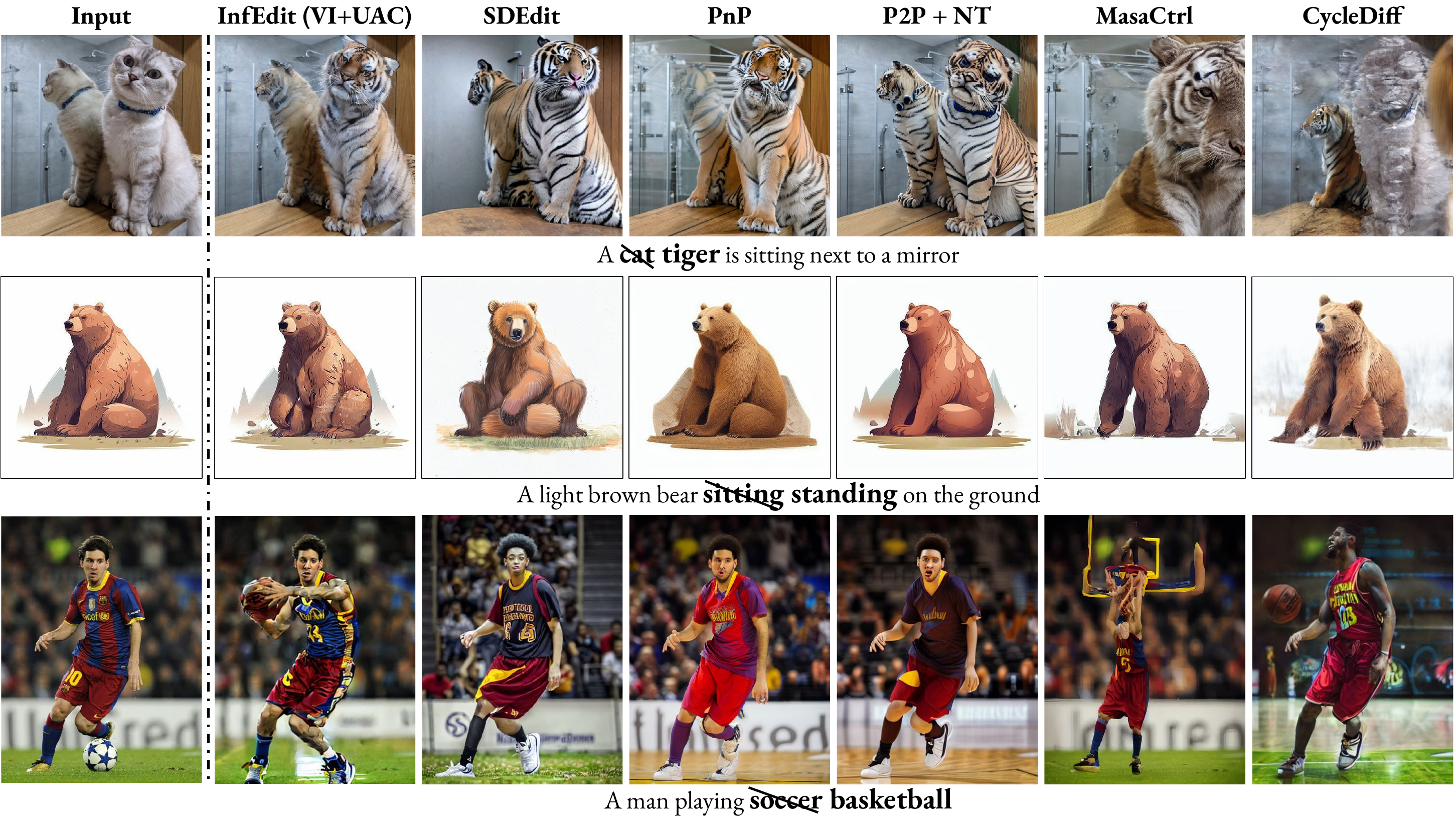}
    \vspace*{-20pt}
    \caption{Qualitative comparisons of InfEdit (VI+UAC) against baselines. InfEdit attains editing goals with the best consistency. \vspace*{-15pt}} 
    \label{fig::qualitative}
\end{figure*}

In this section, we present experiments to demonstrate that \textbf{inversion-free image editing (InfEdit) competes with the effectiveness of inversion-based methods, while also being significantly more efficient.}
Recall that our InfEdit framework adopts Virtual Inversion (VI) derived from DDCM as the sampling framework, and takes any attention control for language-guided editing.
We compare VI against other inversion-based methods on PIE-Bench, with 2 variants of InfEdit (VI+P2P and VI+UAC) for ablation.
The inversion baselines we considered include DDIM~\cite{song2020ddim}, Null-Text (NT)~\cite{mokady2023nti}, Negative Prompt (NP)~\cite{miyake2023npi}, StyleDiffusion (StyleD)~\cite{li2023stylediffusion}, CycleDiffusion (CycleD)~\cite{wu2023cyclediffusion}, and Direct Inversion (DI)~\cite{ju2023direct}.

As depicted in Table~\ref{tab:inversion_based_editing}, InfEdit competes with and often surpasses the effectiveness of inversion-based methods, especially in terms of background consistency.
Additionally, InfEdit is significantly more efficient than inversion-based methods.
On one hand, InfEdit does not require any inversion time. On the other hand, InfEdit generates high-quality target images with much fewer forward steps, and is even compatible with the LCM.
We also present a qualitative example in Figure~\ref{fig::comp-inv}, demonstrating that with the same P2P attention control, VI allows faithful semantic changes as well as better consistency.

It is important to note a fundamental trade-off between reducing image editing distance and improving the faithfulness of image editing. 
StyleD~\cite{li2023stylediffusion}, while demonstrating lower editing distances, exhibits limitations in effective editing, as evidenced by its scores in background preservation and CLIP similarity metrics. 
DI~\cite{ju2023direct} and CycleD~\cite{wu2023cyclediffusion}, surpassing InfEdit in structure distance, often fail to comply with editing instructions, leaving the source image untouched.

\subsection{Attention Control Comparison}
\label{subsec:attention-control}

In this section, we present experiments to demonstrate that \textbf{with unified attention control (UAC), InfEdit establishes state-of-the-art performance in terms of editing quality, consistency, and efficiency.}
We compare UAC with other attention control baselines, especially Prompt-to-Prompt (P2P)~\cite{hertz2023p2p}, Plug-and-Play (PnP)~\cite{tumanyan2023pnp}, and Mutual Self-Attention Control (MasaCtrl)~\cite{cao2023masactrl}.

The comprehensive analysis across 9 distinct categories of editing tasks in PIE-Bench demonstrates the superior performance of InfEdit with UAC as the attention control mechanism, evidenced by enhanced editing quality in Figure~\ref{fig:quality} and improved image consistency in Figure~\ref{fig:consistency}. 
Qualitative comparisons are provided in Figure~\ref{fig::qualitative}, and more results are available in the Appendix.

\subsection{Image-to-Image Translation Tasks}
\label{subsec:i2i-translation}

We further evaluate InfEdit (VI+UAC) in scene-level and object-level I2I translation tasks for more general comparisons.
The baselines we considered include Text2LIVE~\cite{bar2022text2live}, SDEdit~\cite{Meng2021SDEditGI}, CycleD~\cite{wu2023cyclediffusion}, NT~\cite{mokady2023nti}, MasaCtrl~\cite{cao2023masactrl}, as well as the training-based state-of-the-art CycleNet~\cite{xu2023cyclenet}.
As shown in Table~\ref{tab:i2i-compare}, InfEdit strikes an effective balance between translation effects and consistency.
Qualitative examples are shown in Figure~\ref{fig:s2w} and \ref{fig:h2z} in Appendix.

\subsection{Computational Efficiency Ablation}

We finally study the editing efficiency of InfEdit. 
As shown in Table~\ref{tab:inversion_based_editing}, InfEdit significantly outperforms the other baselines in terms of computational efficiency even without applying the LCM.
We perform an ablation study to demonstrate that \textbf{InfEdit gains an advantage through its distinctive compatibility with latent consistency models, facilitating both efficient and high-quality image editing}.
We compare the base diffusion backbones, Stable Diffusion (SD) 1.4~\cite{rombach2022ldm} and Latent Consistency Model (LCM)~\cite{luo2023lcm}, by the CLIP Scores at different forward steps.
Table~\ref{tab:efficiency} (also visualized in Figure~\ref{fig:efficiency}) shows that the LCM significantly outperforms SD 1.4 in editing quality, even with fewer sampling steps. 
While the CLIP Scores of SD incrementally improves from 21.47 to 25.18 as the number of forward steps increases from 2 to 32, LCM can achieve consistently higher CLIP Scores across varying step counts, showing superior speed in image editing.

\begin{table}[!h]
\centering
    \scalebox{0.75}{
    \begin{tabular}{cccccc}
    \toprule
    Method & \multicolumn{5}{c}{\textbf{CLIP Score}} \\ 
    \cmidrule(r){1-1} \cmidrule(r){2-6}
    \textbf{InfEdit (VI+P2P)} & \textbf{2 steps}  & \textbf{4 steps} & \textbf{8 steps} & \textbf{16 steps} & \textbf{32 steps} \\ 
    \cmidrule(r){2-6}
    \textbf{SD 1.4} & 21.47 & 22.17 & 23.16 & 24.45 & 25.18\\
    \textbf{LCM} & \textbf{25.76} & \textbf{25.33} & \textbf{24.76} &	\textbf{25.08} & \textbf{26.68} \\
    \bottomrule
    \end{tabular}}
    \vspace*{-5pt}
    \caption{The use of LCM backbone in InfEdit allows a high CLIP Score even with fewer steps. \vspace*{-18pt}}
    \label{tab:efficiency}
\end{table}

\begin{table*}[htbp]
\small
\centering
  \renewcommand\arraystretch{1.0}
\setlength{\tabcolsep}{0.7mm}{
\begin{threeparttable}
\begin{tabular}{crccccrcccc}
\toprule
\textbf{Task} & \multicolumn{5}{c}{\textbf{Summer$\leftrightarrow$Winter ($512\times512$)}} & \multicolumn{5}{c}{\textbf{Horse$\leftrightarrow$Zebra ($512\times512$)}} \\
\cmidrule(r){1-1} \cmidrule(r){2-6} \cmidrule(r){7-11} 
\textbf{Method} & \textbf{FID} $\downarrow$ & \textbf{CLIP Sim} $\uparrow$ & \textbf{LPIPS} $\downarrow$ & \textbf{PSNR} $\uparrow$ & \textbf{SSIM} $\uparrow$ & \textbf{FID} $\downarrow$ & \textbf{CLIP Sim} $\uparrow$ & \textbf{LPIPS} $\downarrow$ & \textbf{PSNR} $\uparrow$ & \textbf{SSIM} $\uparrow$ \\
\cmidrule(r){1-1} \cmidrule(r){2-6} \cmidrule(r){7-11} 
\textcolor{gray}{\textbf{CycleNet}} & \textcolor{gray}{79.79} & \textcolor{gray}{24.12} & \textcolor{gray}{0.15} & \textcolor{gray}{25.88} & \textcolor{gray}{0.69} & \textcolor{gray}{76.83} & \textcolor{gray}{25.27} & \textcolor{gray}{0.08} & \textcolor{gray}{26.21} & \textcolor{gray}{0.74} \\
\cmidrule(r){1-1} \cmidrule(r){2-6} \cmidrule(r){7-11} 
\textbf{Text2LIVE} & 86.12 & \textbf{25.98} & 0.27 & 16.83 & 0.68 & 103.14 & 31.55 & \textbf{0.16} & 20.98 & \textbf{0.81} \\
\textbf{SDEdit} & 90.51 & 23.26 & 0.30 & 18.59 & 0.43 & 63.04 & 27.97 & 0.33 & 18.49 & 0.44 \\
\textbf{CycleD} & 84.52 & 24.40 & 0.24 & 21.66 & 0.68 & \textbf{41.17} & \textbf{29.09} & 0.29 & 19.41 & 0.61 \\
\textbf{NT+P2P} & 92.65 & 24.82 & 0.24 & 20.19 & 0.66 & 106.83 & 26.57 & 0.21 & 21.45 & 0.66 \\
\textbf{MasaCtrl} & 114.83 & 17.11 & 0.37 & 14.66 & 0.43 & 239.61 & 21.15 & 0.41 & 16.31 & 0.37 \\
\cmidrule(r){1-1} \cmidrule(r){2-6} \cmidrule(r){7-11} 
\rowcolor[HTML]{C9DAF8}
\textbf{InfEdit} & \textbf{75.63} & 23.07 & \textbf{0.18} & \textbf{21.99} & \textbf{0.68} & 61.81 & 28.16 & \textbf{0.16} & \textbf{21.80} & 0.72 \\
\bottomrule
\end{tabular}
   \footnotesize
    \end{threeparttable}}
    \vspace*{-10pt}
    \caption{Image2Image translation comparison. InfEdit methods achieve a favorable balance between consistency and translation quality. \vspace*{-15pt}}
    \label{tab:i2i-compare}
\end{table*}

\subsection{Connecting InfEdit to LLMs}

Based on the capability of InfEdit to edit images with natural language, we can further leverage large language models (LLMs) to follow the image editing instructions. 
Through our experiments, we have validated the feasibility of prompting GPT-4~\cite{openai2023gpt4} to break down editing instructions into adequate source and target prompts for InfEdit. 
This allows users to give natural language instructions to control image editing, which improves the user experience.
A Gradio demo is available on our project page.

\section{Related Work}

\vspace{-5pt}
\paragraph{Image Manipulation with Diffusion Models}
Diffusion models (DMs) have achieved notable success in image generation~\cite{song2020ddim, ho2020ddpm}, with large-scale models pre-trained on text-to-image tasks~\cite{Nichol2021GLIDETP,rombach2022ldm}. 
These models are increasingly adopted for image manipulation tasks, where DMs are augmented with additional conditions like text prompts~\cite{Ramesh2022HierarchicalTI} or images~\cite{zhang2023adding,xu2023cyclenet} to generate the target image.
The source image information is usually integrated into DMs through an inversion process~\cite{song2020ddim} or via a side network~\cite{zhang2023adding, mou2023t2i}. 
Additionally, mask-based methods have been proposed, utilizing either user-prompted or automatically generated masks~\cite{nichol2022glide,couairon2023diffedit, avrahami2023spatext}, or augmentation layers \cite{bar2022text2live}, to facilitate more controlled and precise image manipulations. 
To enhance the consistency and quality of image edits, several techniques have been developed. 
Among these, attention control mechanisms~\cite{hertz2023p2p,tumanyan2023pnp,cao2023masactrl} have emerged as a promising direction, especially when they are paired with inversion methods~\cite{mokady2023nti,miyake2023npi,ju2023direct}.

\vspace{-5pt}
\paragraph{Inversion in Diffusion Models}
In DMs, real image editing methods usually rely on the inversion process, which produces a latent representation that can reconstruct the image through the generative process. 
Initially, SDEdits~\cite{Meng2021SDEditGI} was proposed, which adds random Gaussian noise to the source image as input but suffers from reconstruction quality. 
DDIM inversion~\cite{hertz2023p2p} was then introduced for its deterministic mapping from latent space to image but is prone to errors accumulated in its multiple-step inversion process.
Null-text inversion~\cite{mokady2023nti} used a null-text prompt for pivot tuning, improving real image editing but was time-consuming and not fully accurate. 
Negative prompt inversion~\cite{miyake2023npi} accelerated the inversion process by approximating the DDIM inversion, while sacrificing the reconstruction quality.
CycleDiffusion~\cite{wu2023cyclediffusion}, Editing-friendly Inversion~\cite{hubermanspiegelglas2023edit} and Direct Inversion~\cite{ju2023direct} use source latents from each inversion step as reference for editing the target branch. 
However, these methods still struggle with the cumulative errors typical of the inversion process and tend to be slower overall due to the inherent need for inversion.

\vspace{-5pt}
\paragraph{Attention-Control for Image Editing} 
In the realm of zero-shot image editing, attention control becomes a pivotal technique to preserve consistency while manipulating visual content. 
Recent works like Prompt-to-Prompt (P2P)~\cite{hertz2023p2p} and Plug-and-Play (PnP)~\cite{tumanyan2023pnp} have contributed significantly to this field by replacing cross-attention and self-attention maps to maintain the original image layout and spatial information, thereby preserving the consistency during editing. 
In contrast, MasaCtrl~\cite{cao2023masactrl} offers an alternative approach that enables the modification of layout and spatial attributes while safeguarding the semantic content inherent in the image, which addresses the limitation of conflating spatial edits with semantic preservation. 
P2Plus~\cite{li2023stylediffusion} extends the prompt-to-prompt paradigm by applying edits to both the text-conditional and unconditional branches during classifier-free guidance~\cite{ho2021classifier}, thus offering a more comprehensive editing framework.

\section{Conclusion}
Recent advancements in inversion-based editing notwithstanding, text-guided image manipulation using diffusion models continues to be a challenge. 
The main challenges involve 1) the lengthy inversion process; 2) difficulties in maintaining both consistency and accuracy; 3) incompatibility with the efficient consistency sampling methods of consistency models. 
In response, we questioned whether it's possible to bypass the inversion process in editing. 
Our findings reveal that with a known initial sample $z_0$, a specific variance schedule $\sigma$ can simplify the denoising step to a form akin to multi-step consistency sampling. 
This led to the development of the Denoising Diffusion Consistent Model (DDCM), which effectively introduces a virtual inversion strategy that eliminates the need for explicit inversion during sampling. 
Moreover, we present the Unified Attention Control (UAC) mechanisms as a tuning-free framework for text-guided editing. 
This integration forms the basis of our inversion-free editing approach, InfEdit, which facilitates consistent and accurate editing across both rigid and non-rigid semantic transformations. 
InfEdit is adept at handling complex modifications without compromising the image's integrity or requiring explicit inversion. 
Extensive experiments demonstrate its robust performance across a range of editing tasks and that it maintains a smooth workflow, completing tasks in under 3 seconds on a single A40.
InfEdit unleashes the potential for real-time image editing applications.

\section*{Ethics Statement}
While InfEdit offers promising advancements in image editing, it is crucial to consider its broader ethical, legal, and societal implications. 

\vspace{-5pt}
\paragraph{Copyright Infringement.}
As an advanced image editing tool, InfEdit could be used to modify and repurpose artists' original works, raising concerns over copyright violations. 
It's vital for practitioners to respect the rights of creators and maintain the integrity of the creative economy, ensuring adherence to licensing and copyright laws.

\vspace{-5pt}
\paragraph{Deceptive Misuse.}
If exploited by nefarious entities, InfEdit's capability to generate convincing image alterations could be used for misinformation, fraud, or identity theft. 
This necessitates responsible user guidelines and strong security protocols to prevent such misuse and safeguard against security threats.

\vspace{-5pt}
\paragraph{Bias and Fairness.}
Furthermore, InfEdit builds upon pre-trained latent diffusion models and latent consistency models, which might carry inherent biases, leading to potential fairness issues. 
While the method is algorithmic and not pre-trained on large web-scale datasets, it's important to recognize and mitigate any encoded biases in these pre-trained backbones to ensure fairness and ethical use.

\vspace{5pt}
\noindent
By proactively addressing these concerns, we can leverage InfEdit's capabilities responsibly, prioritizing ethical considerations, legal compliance, and the welfare of society. This approach is essential for advancing technology while safeguarding our community's values and trust.

{
    \small
    \bibliographystyle{ieeenat_fullname}
    \bibliography{main}
}




\begin{figure*}[!htp]
    \centering
    \includegraphics[width=0.87\linewidth]{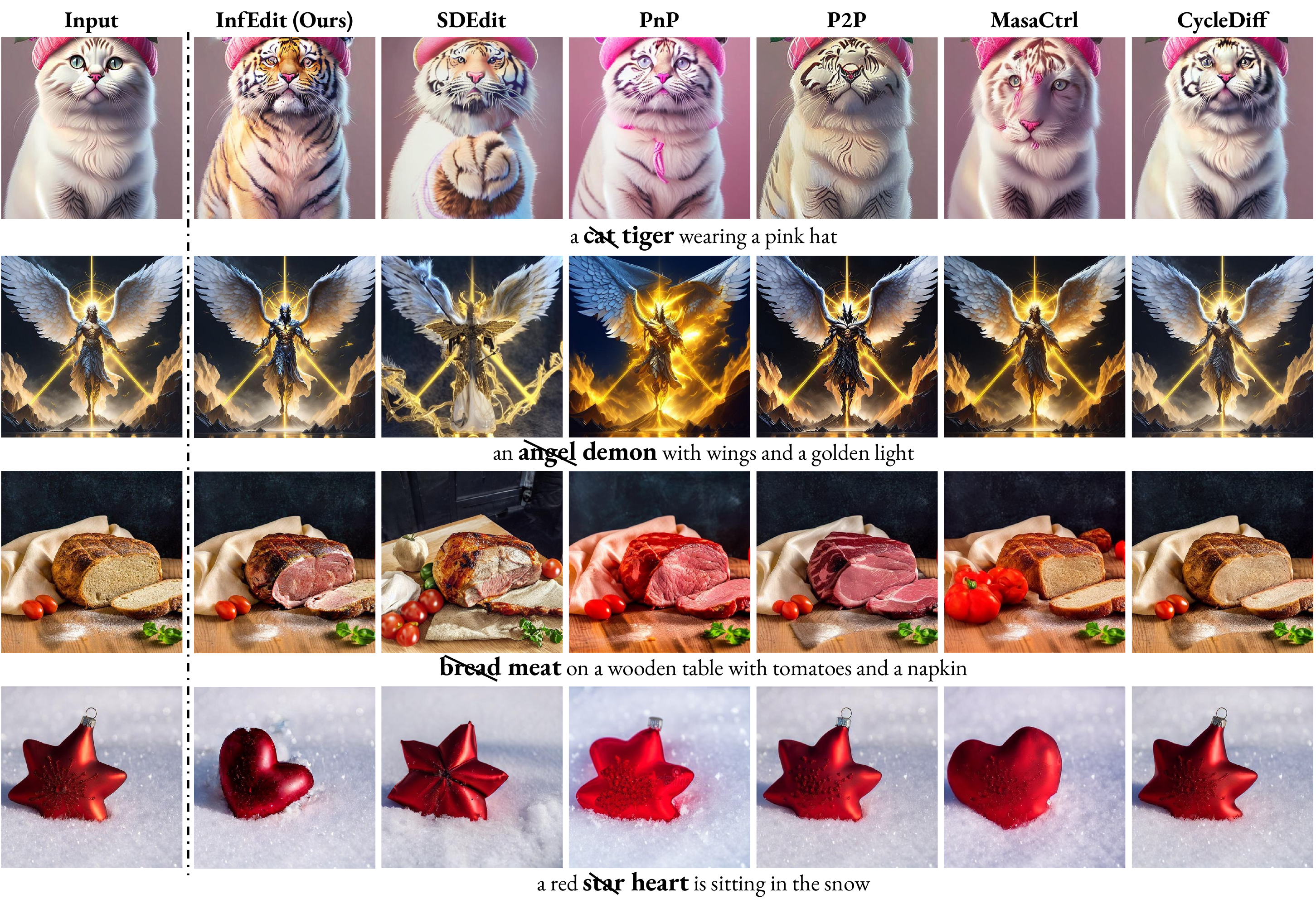}
    \vspace{-10pt}
    \caption{Additional comparison on changing object tasks.} 
\end{figure*}
\begin{figure*}[!htp]
    \centering
    \includegraphics[width=0.87\linewidth]{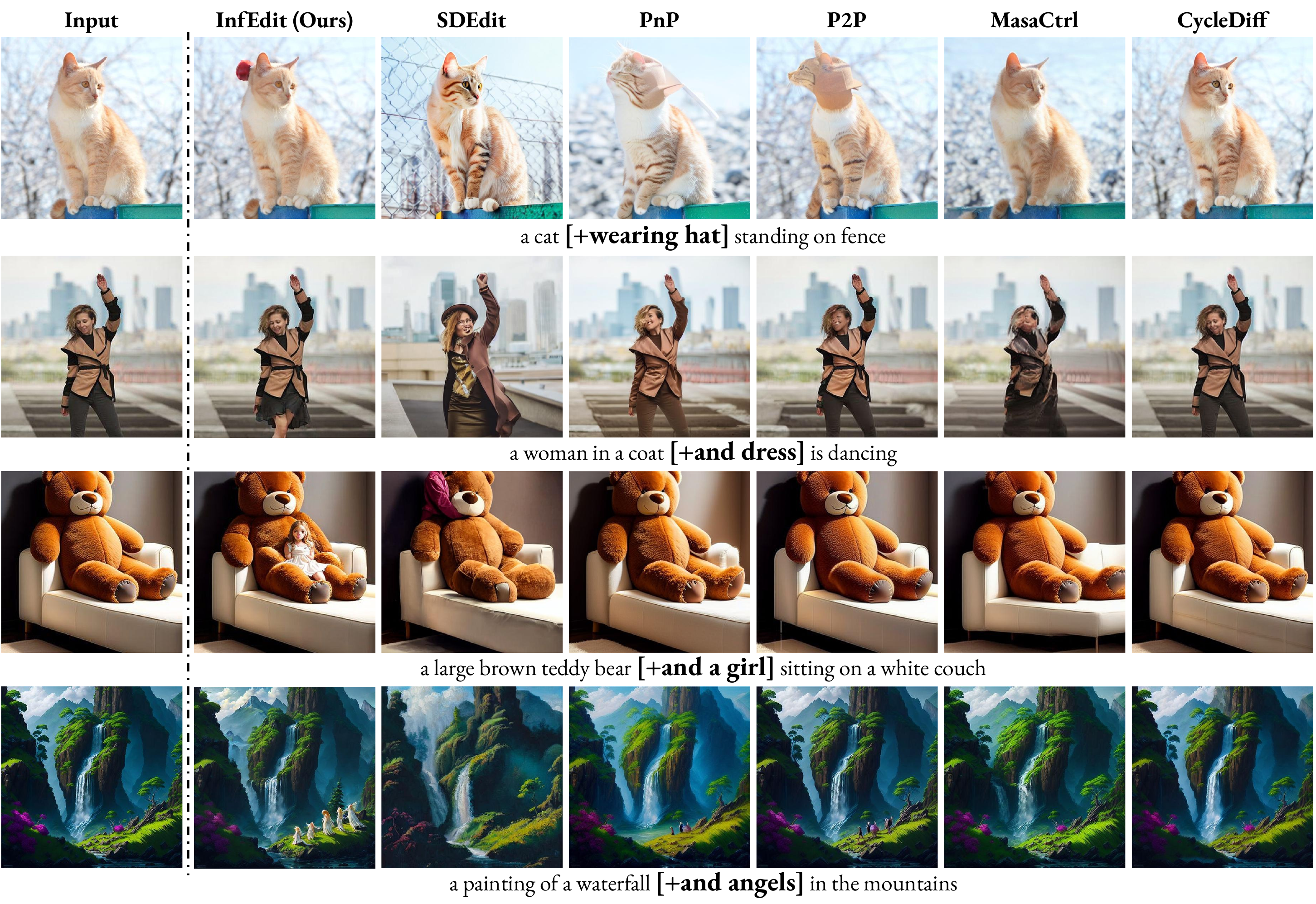}
    \vspace{-10pt}
    \caption{Additional comparison on adding object tasks.} 
\end{figure*}
\begin{figure*}[!htp]
    \centering
    \includegraphics[width=0.87\linewidth]{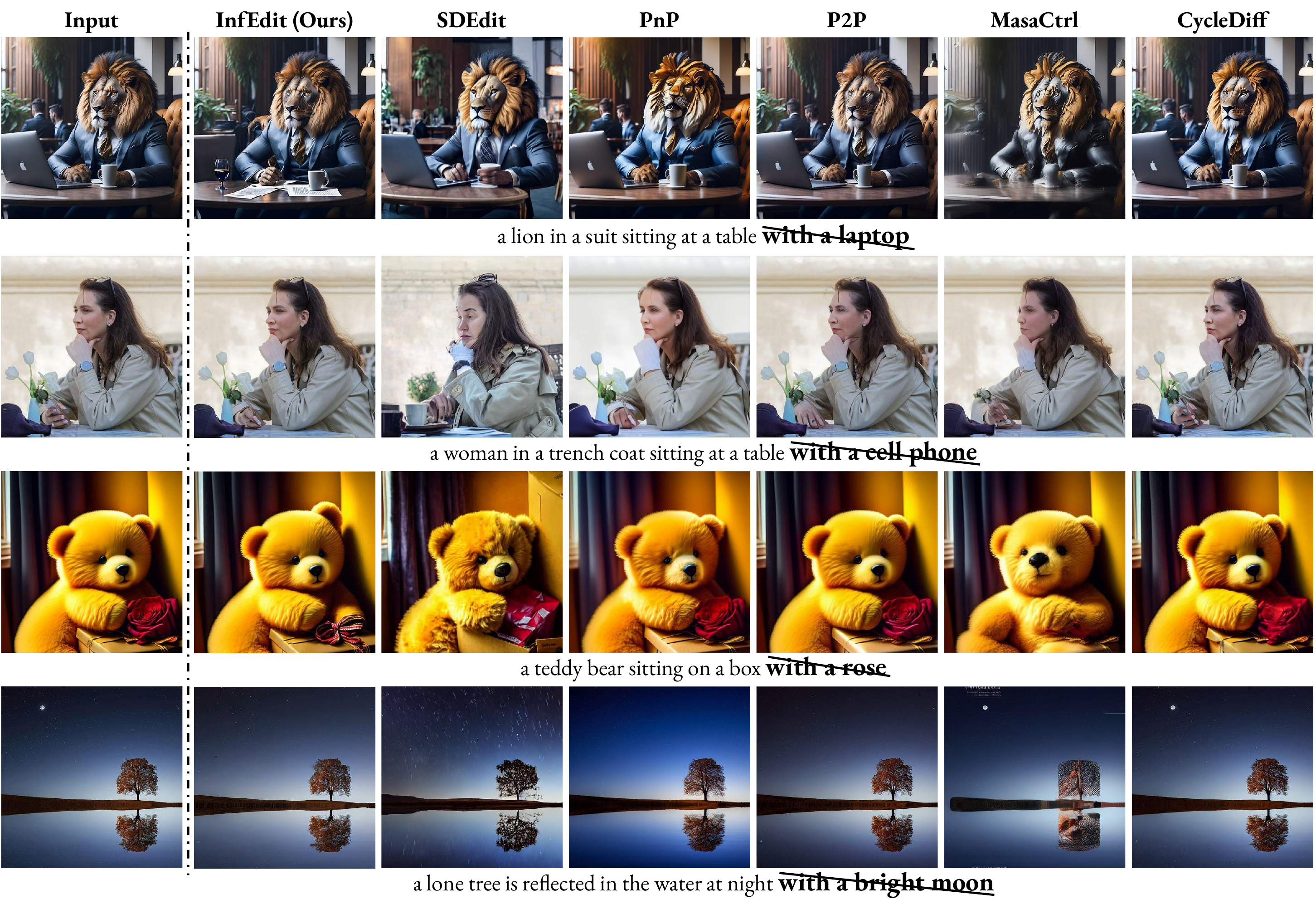}
    \vspace{-10pt}
    \caption{Additional comparison on deleting object tasks.} 
\end{figure*}
\begin{figure*}[!htp]
    \centering
    \includegraphics[width=0.87\linewidth]{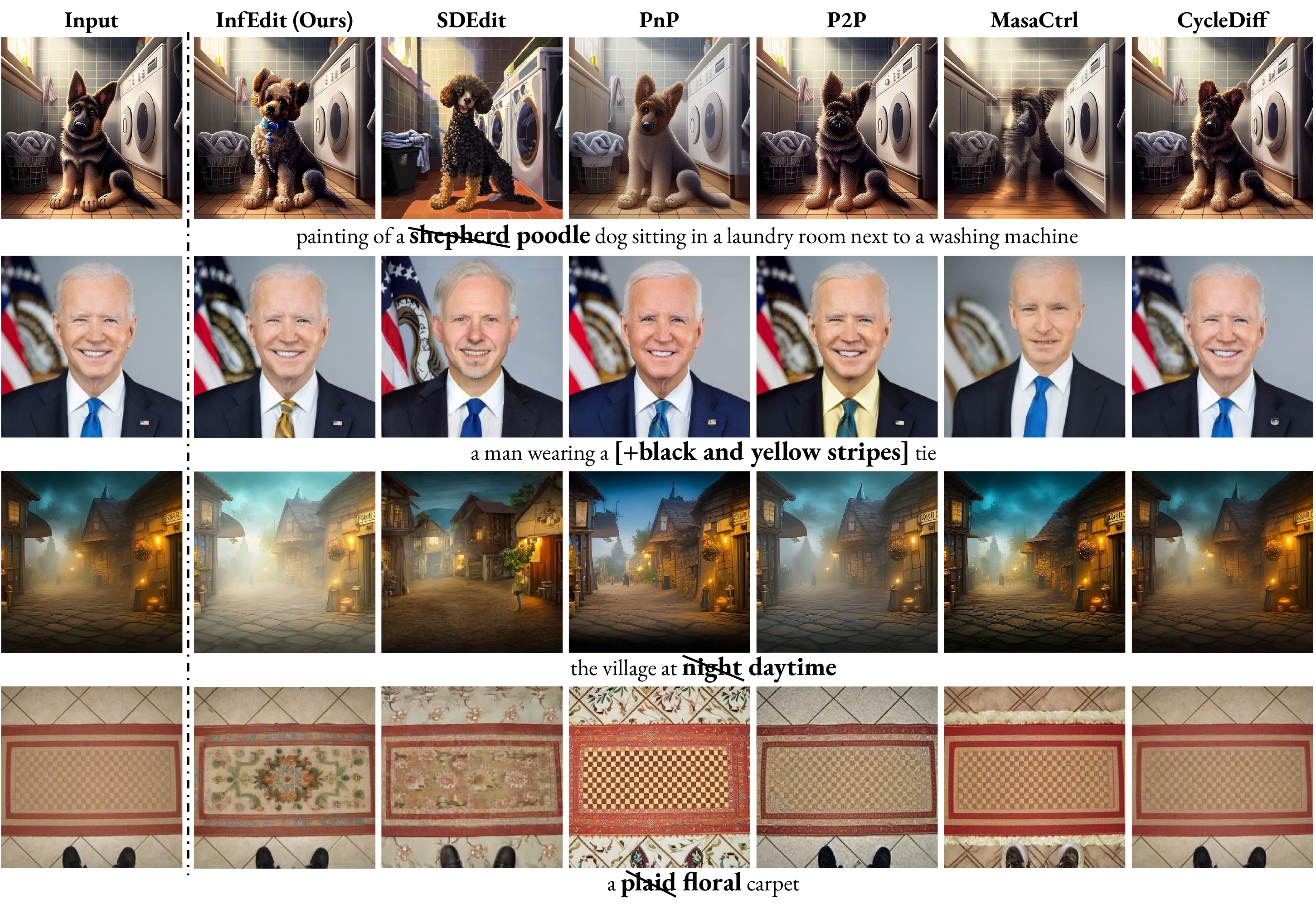}
    \vspace{-10pt}
    \caption{Additional comparison on changing content tasks.} 
\end{figure*}
\begin{figure*}[!htp]
    \centering
    \includegraphics[width=0.87\linewidth]{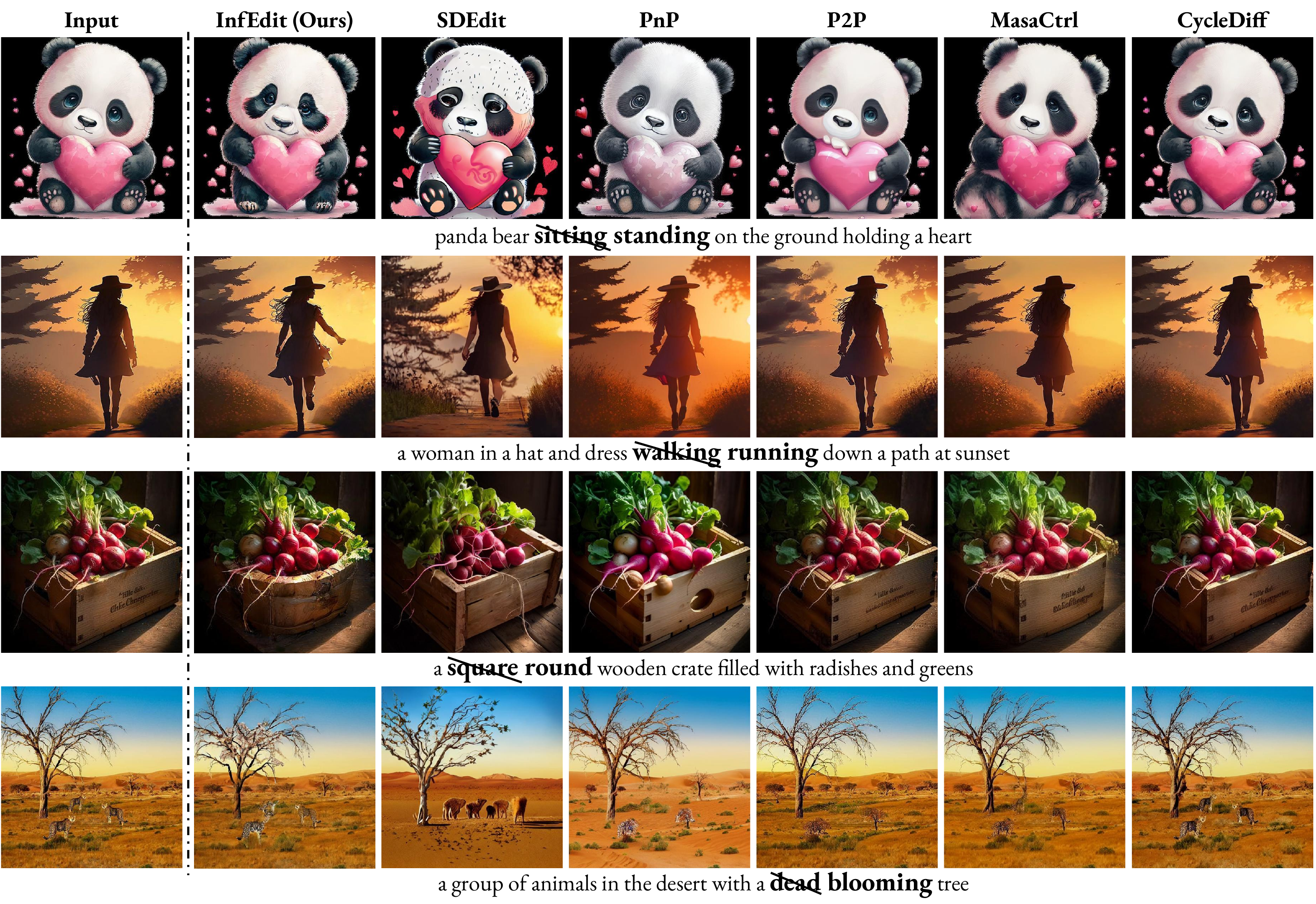}
    \vspace{-10pt}
    \caption{Additional comparison on changing pose tasks.} 
\end{figure*}
\begin{figure*}[!htp]
    \centering
    \includegraphics[width=0.87\linewidth]{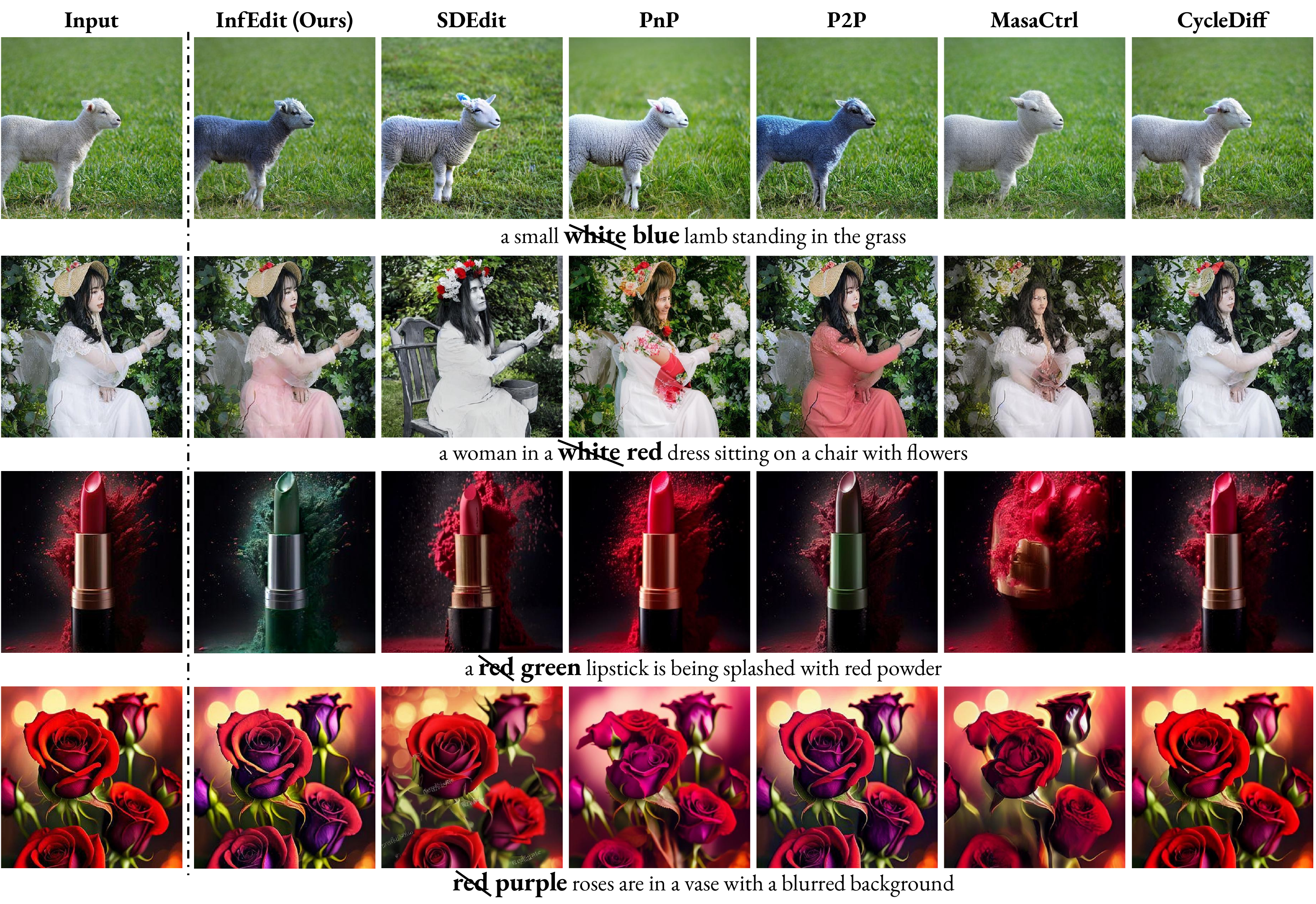}
    \vspace{-10pt}
    \caption{Additional comparison on changing color tasks.} 
\end{figure*}
\begin{figure*}[!htp]
    \centering
    \includegraphics[width=0.87\linewidth]{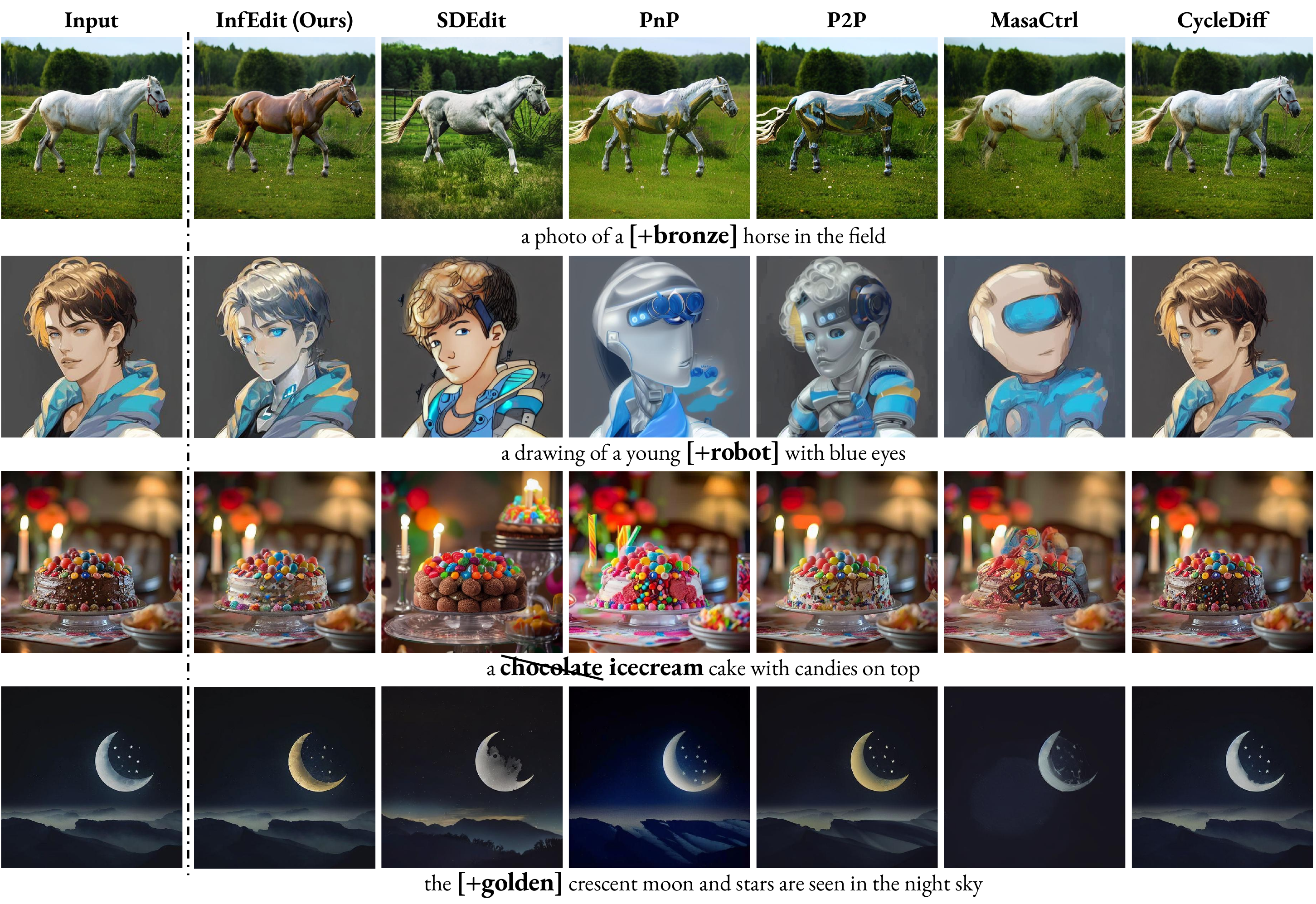}
    \vspace{-10pt}
    \caption{Additional comparison on changing material tasks.} 
\end{figure*}
\begin{figure*}[!htp]
    \centering
    \includegraphics[width=0.87\linewidth]{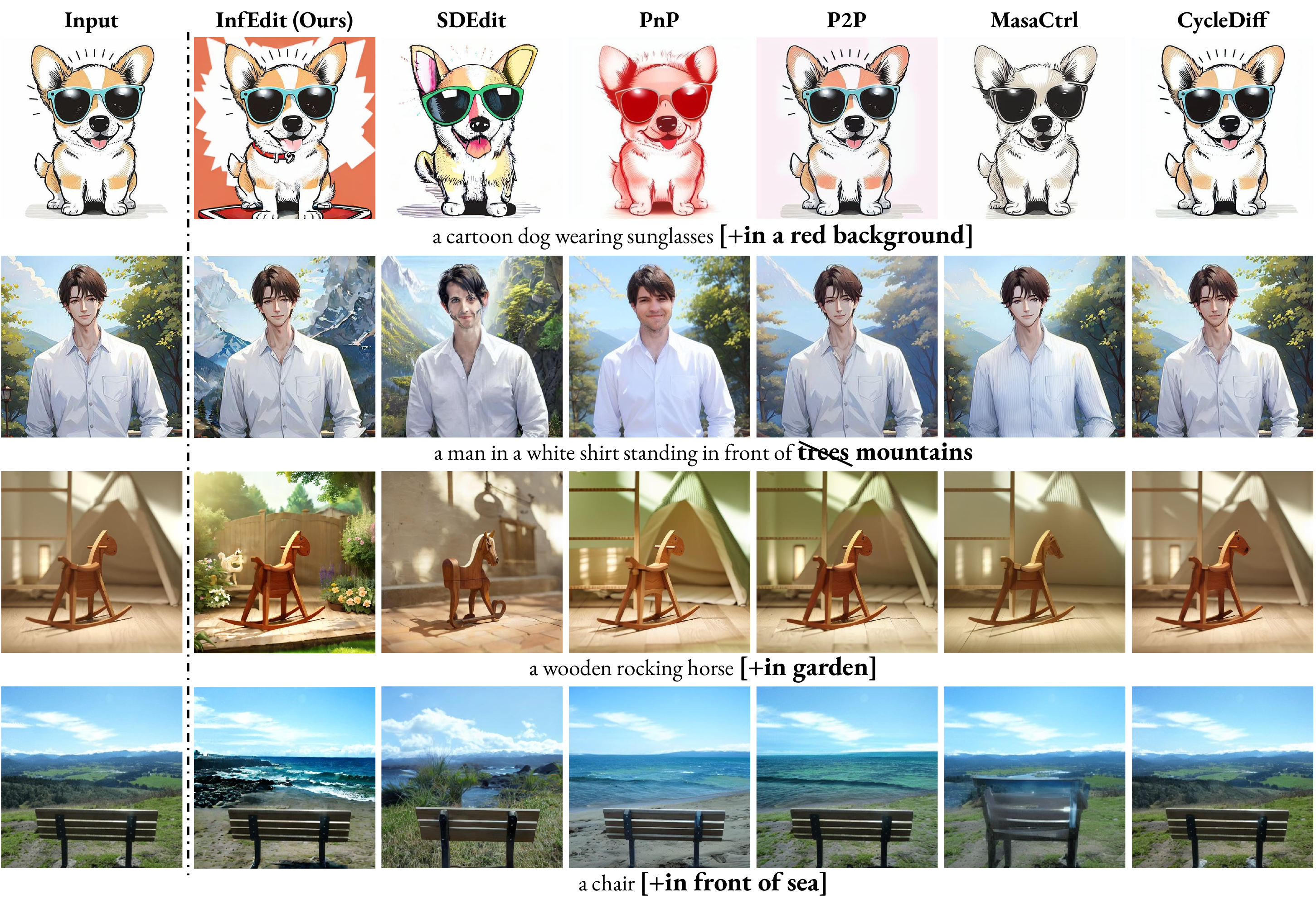}
    \vspace{-10pt}
    \caption{Additional comparison on changing background tasks.} 
\end{figure*}
\begin{figure*}[!htp]
    \centering
    \includegraphics[width=0.87\linewidth]{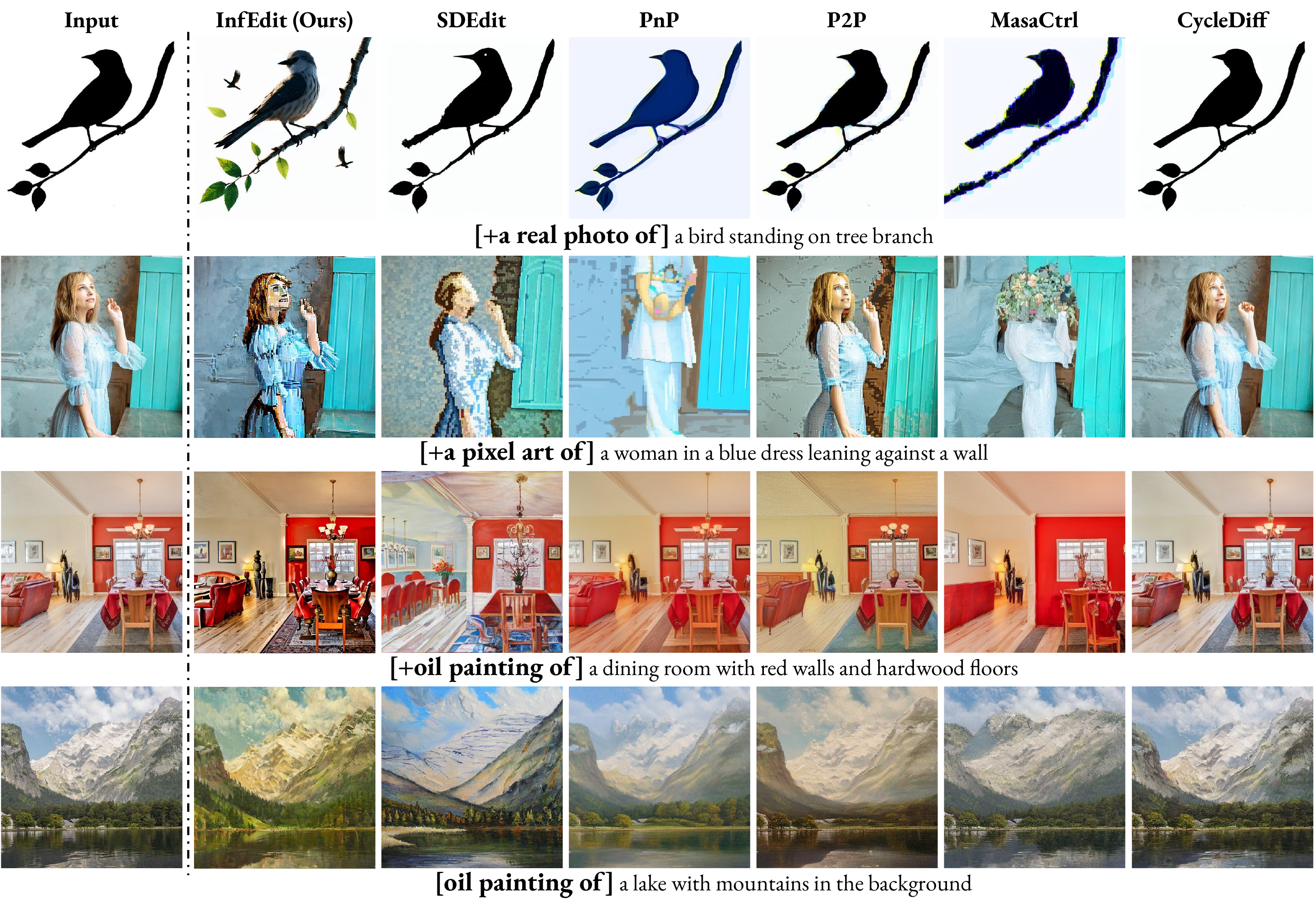}
    \vspace{-10pt}
    \caption{Additional comparison on changing style tasks.} 
\end{figure*}
\clearpage

\begin{figure*}[!htp]
    \centering
    \includegraphics[width=1.0\linewidth]{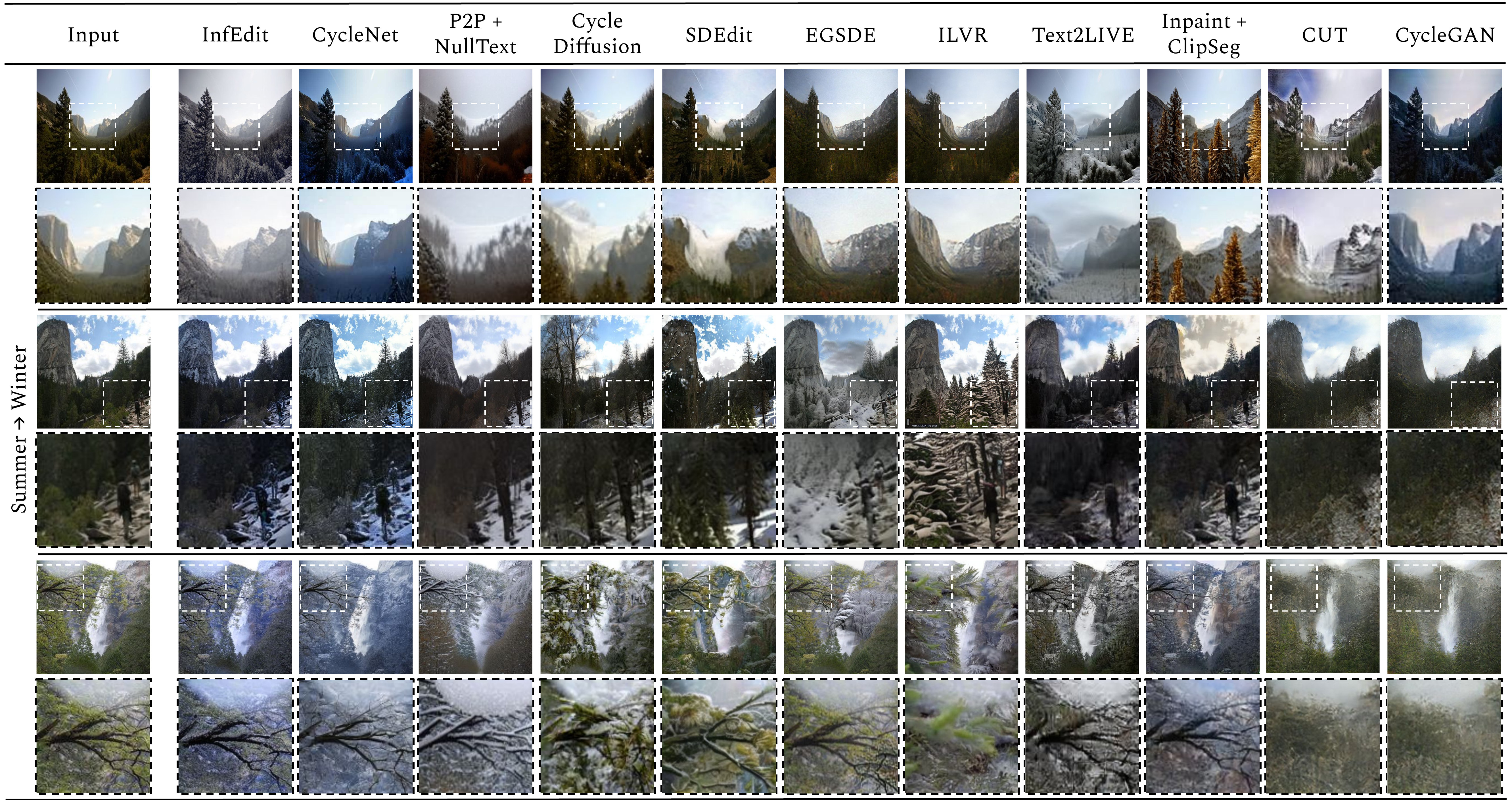}
    \caption{Qualitative comparison of InfEdit on Summer2Winter task with other baselines.} 
    \label{fig:s2w}
\end{figure*}

\begin{figure*}[!htp]
    \centering
    \includegraphics[width=1.0\linewidth]{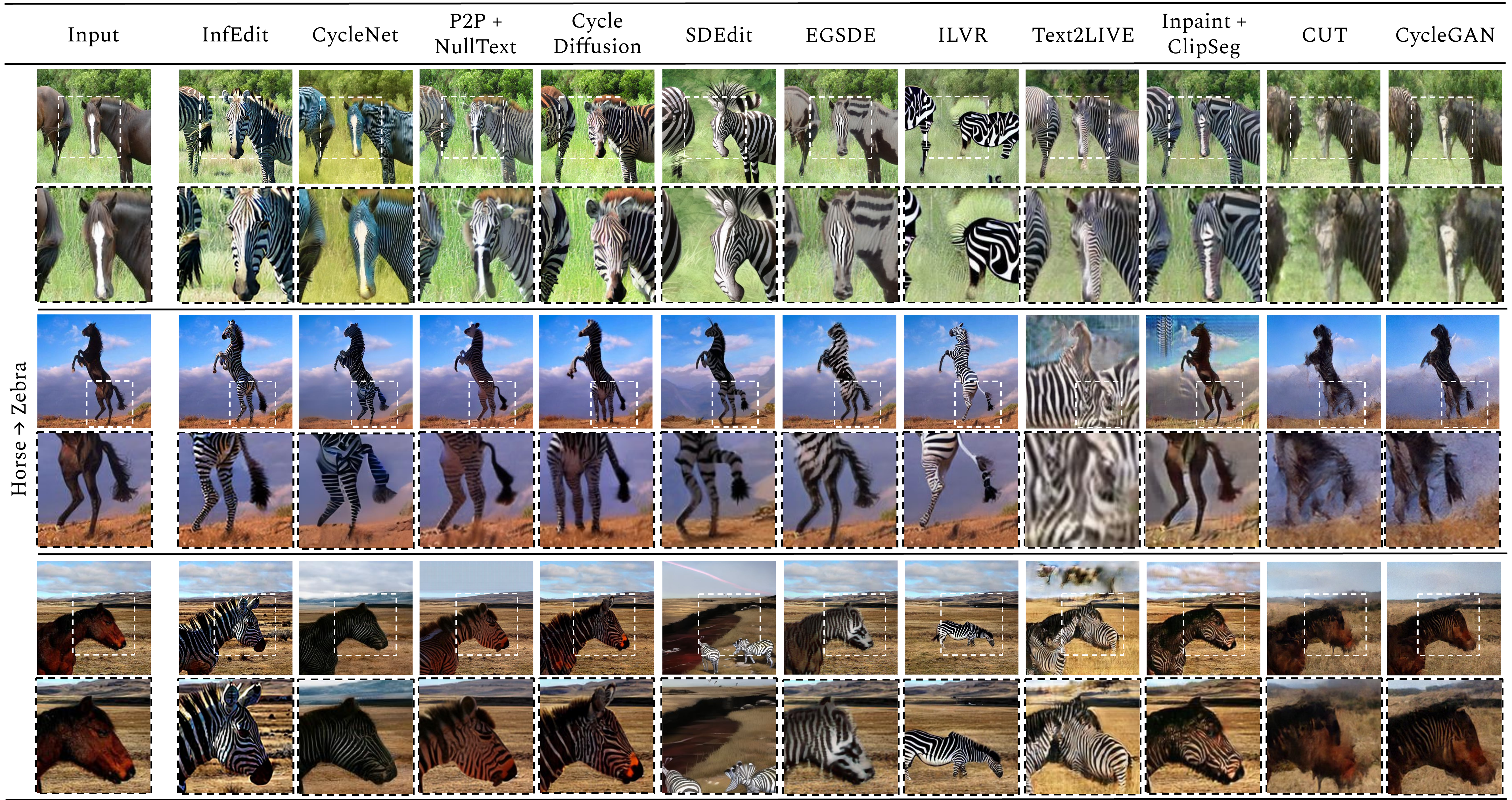}
    \caption{Qualitative comparison of InfEdit on Horse2Zebra task with other baselines.} 
    \label{fig:h2z}
\end{figure*}
\clearpage

\begin{figure*}[!htp]
    \centering
    \includegraphics[width=1.0\linewidth]{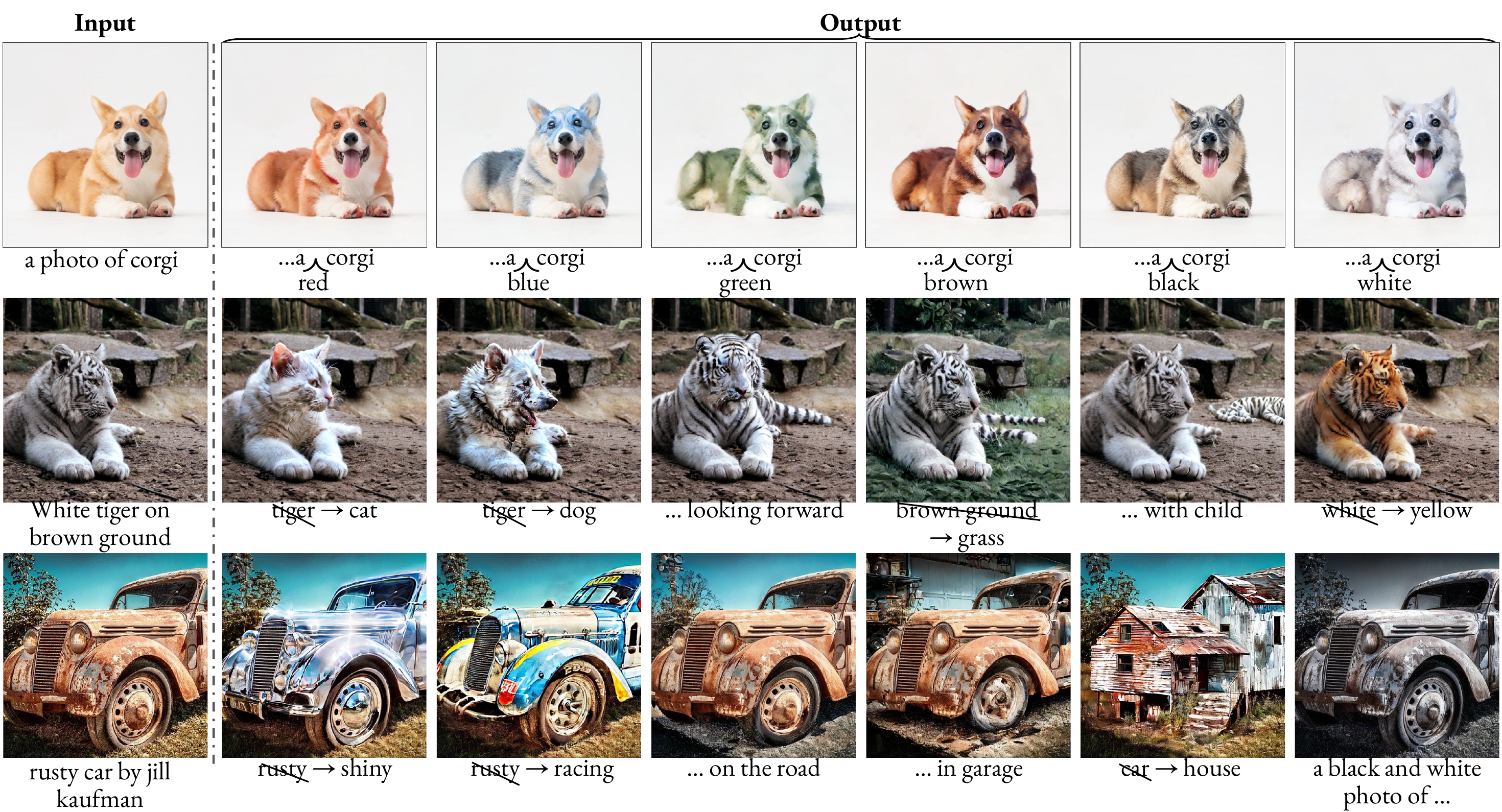}
    \vspace{-10pt}
    \caption{Additional results of InfEdit in various complex image editing tasks.} 
\end{figure*}

\begin{figure*}[!htp]
    \centering
    \includegraphics[width=1.0\linewidth]{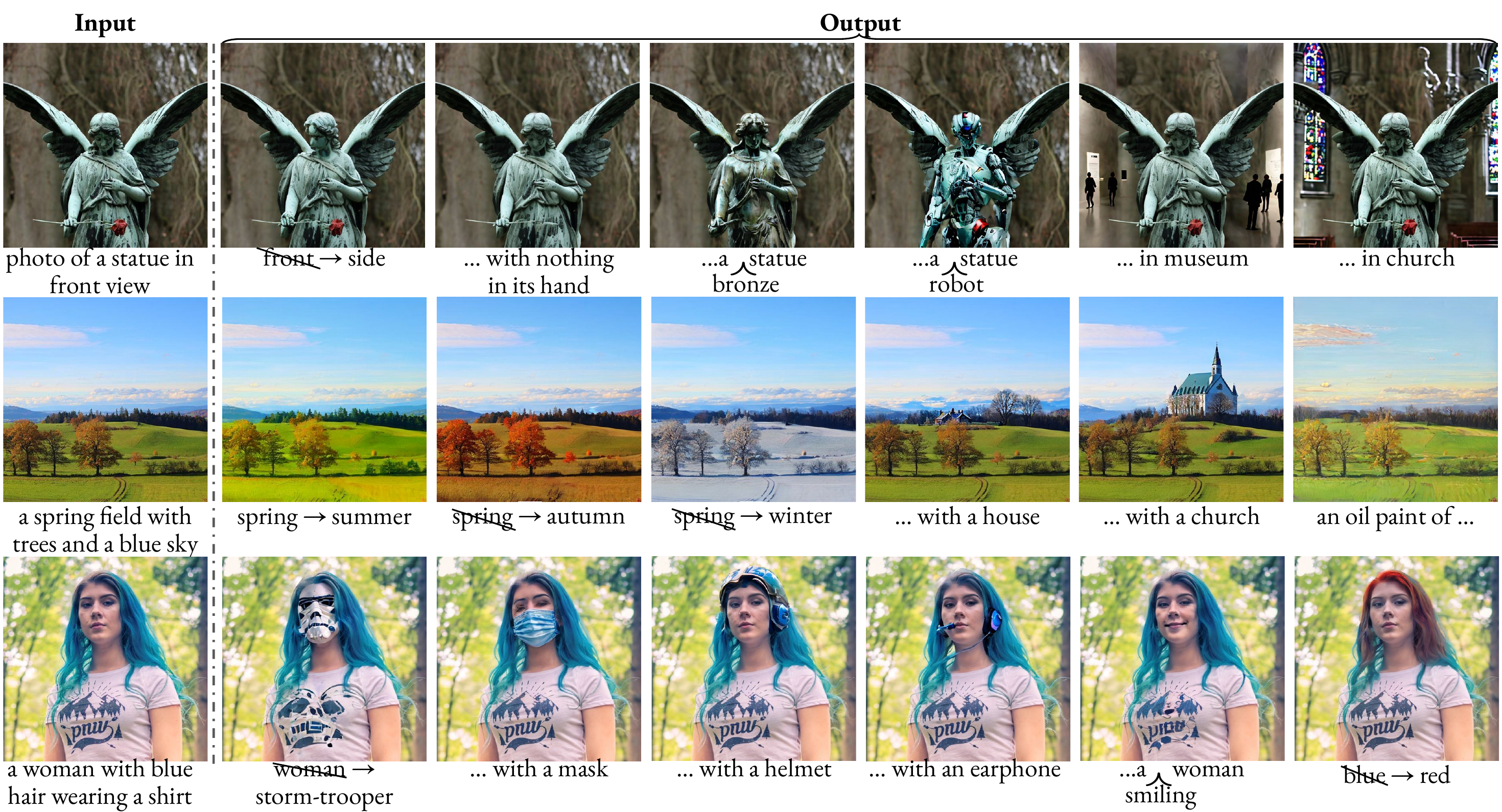}
    \vspace{-10pt}
    \caption{Additional results of InfEdit in various complex image editing tasks.} 
\end{figure*}

\begin{figure*}[!htp]
    \centering
    \includegraphics[width=1.0\linewidth]{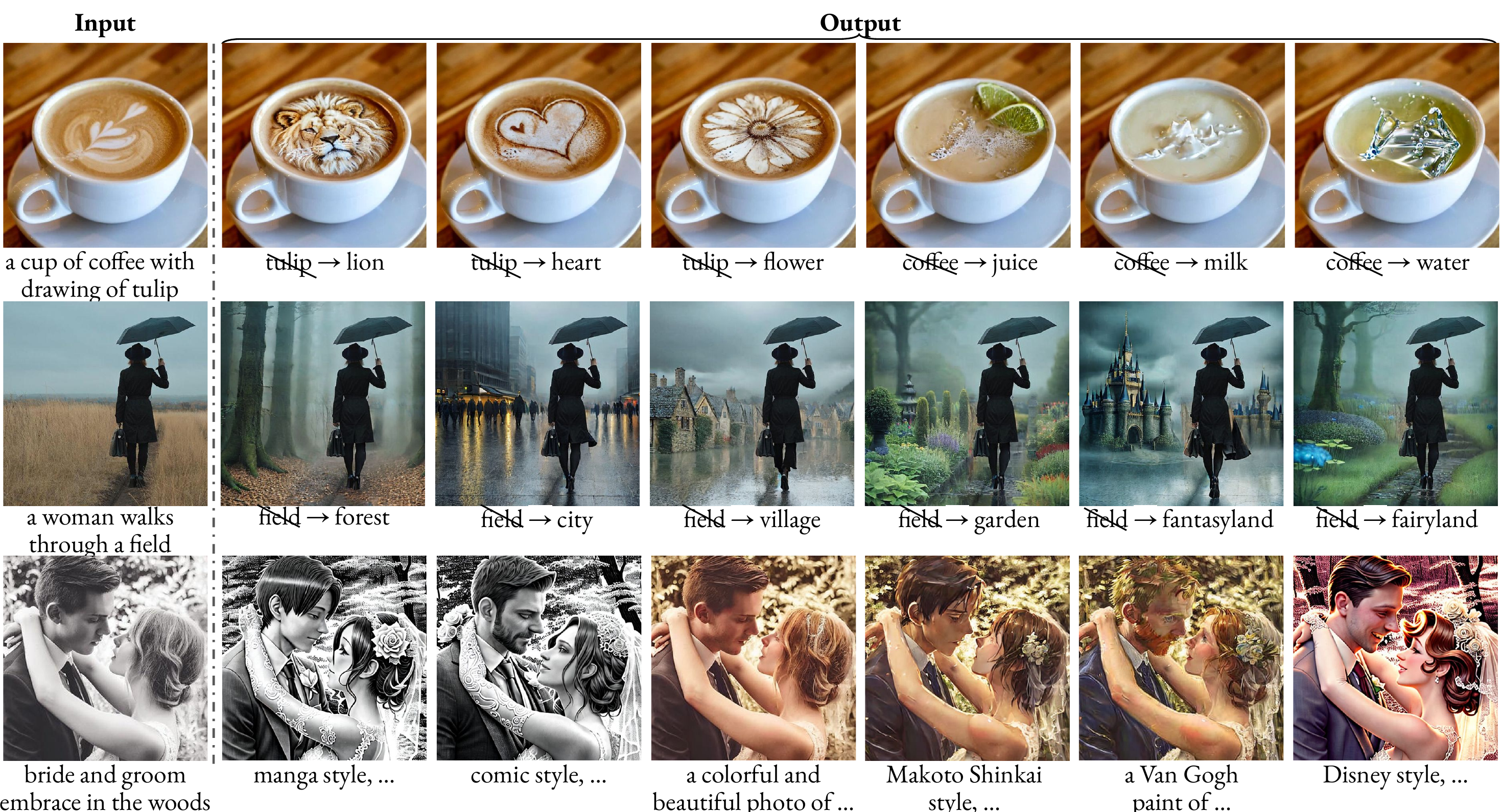}
    \vspace{-10pt}
    \caption{Additional results of InfEdit in various complex image editing tasks.} 
\end{figure*}

\clearpage

\begin{figure*}[!htp]
    \centering
    \includegraphics[width=1.0\linewidth]{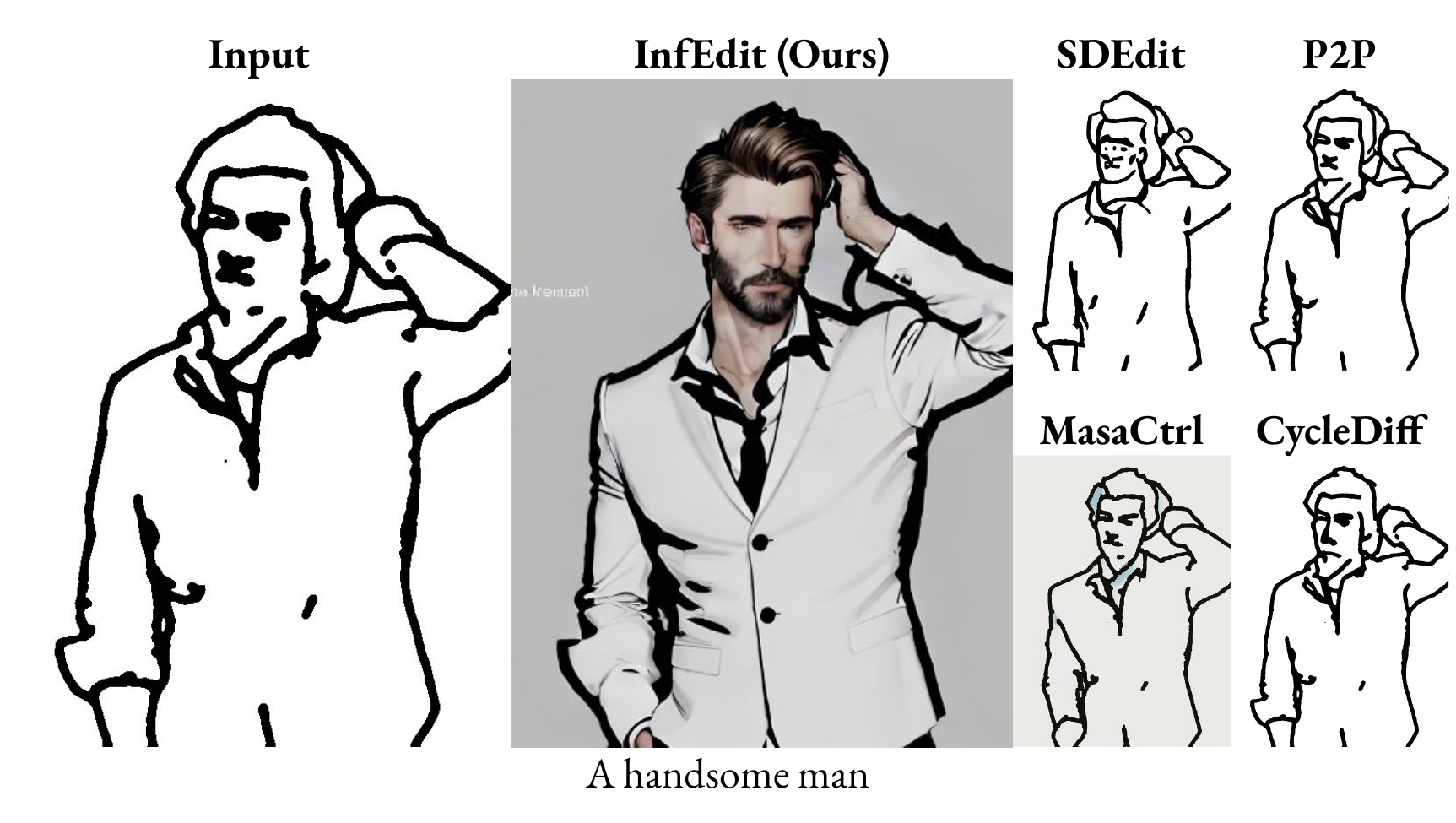}
    \vspace{-10pt}
    \caption{Additional results of InfEdit compared with other method.} 
\end{figure*}

\begin{figure*}[!htp]
    \centering
    \includegraphics[width=1.0\linewidth]{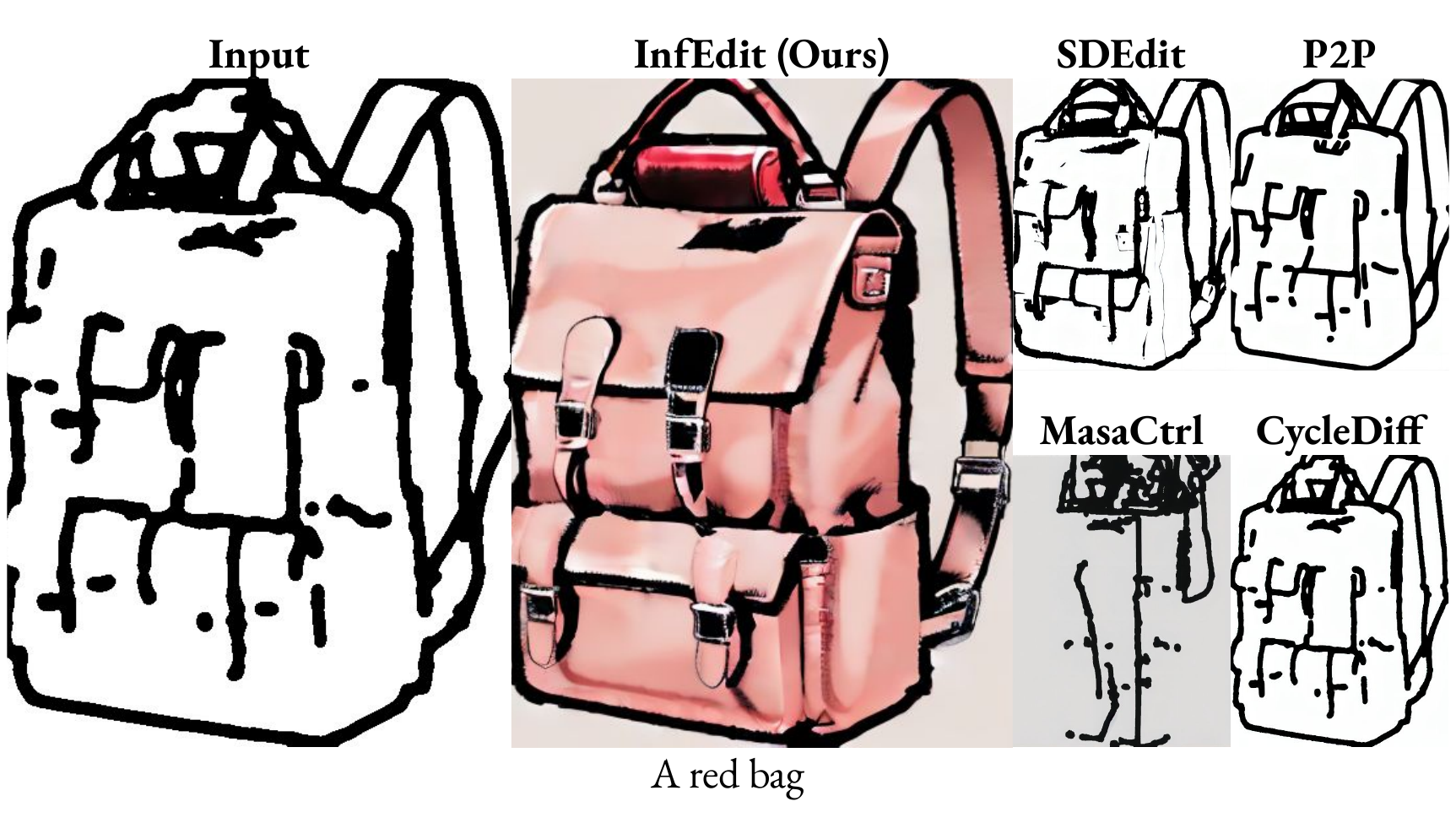}
    \vspace{-10pt}
    \caption{Additional results of InfEdit compared with other method.} 
\end{figure*}

\begin{figure*}[!htp]
    \centering
    \includegraphics[width=1.0\linewidth]{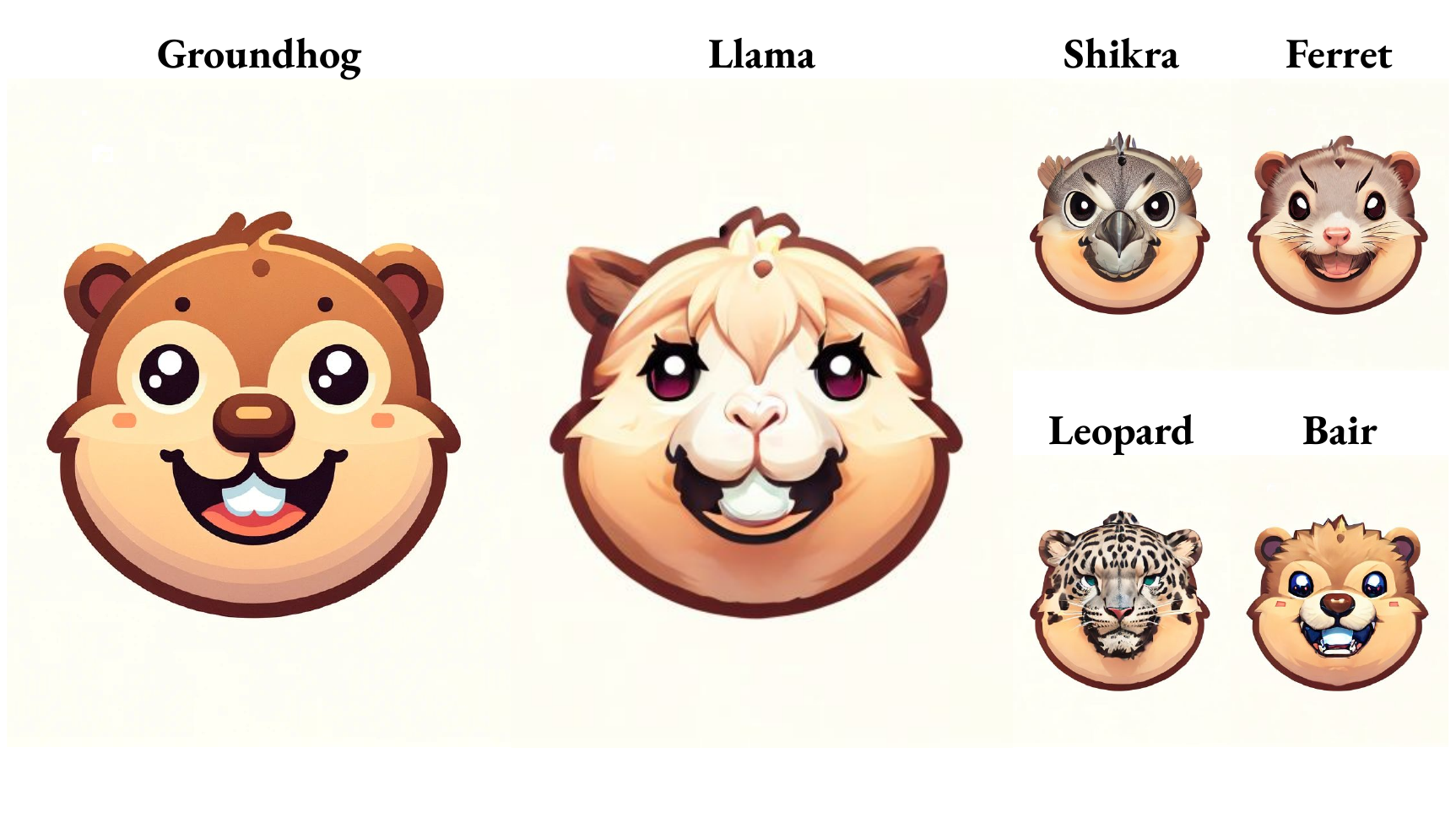}
    \vspace{-10pt}
    \caption{Additional results of InfEdit in multi-modal editing.} 
\end{figure*}

\begin{figure*}[!htp]
    \centering
    \includegraphics[width=1.0\linewidth]{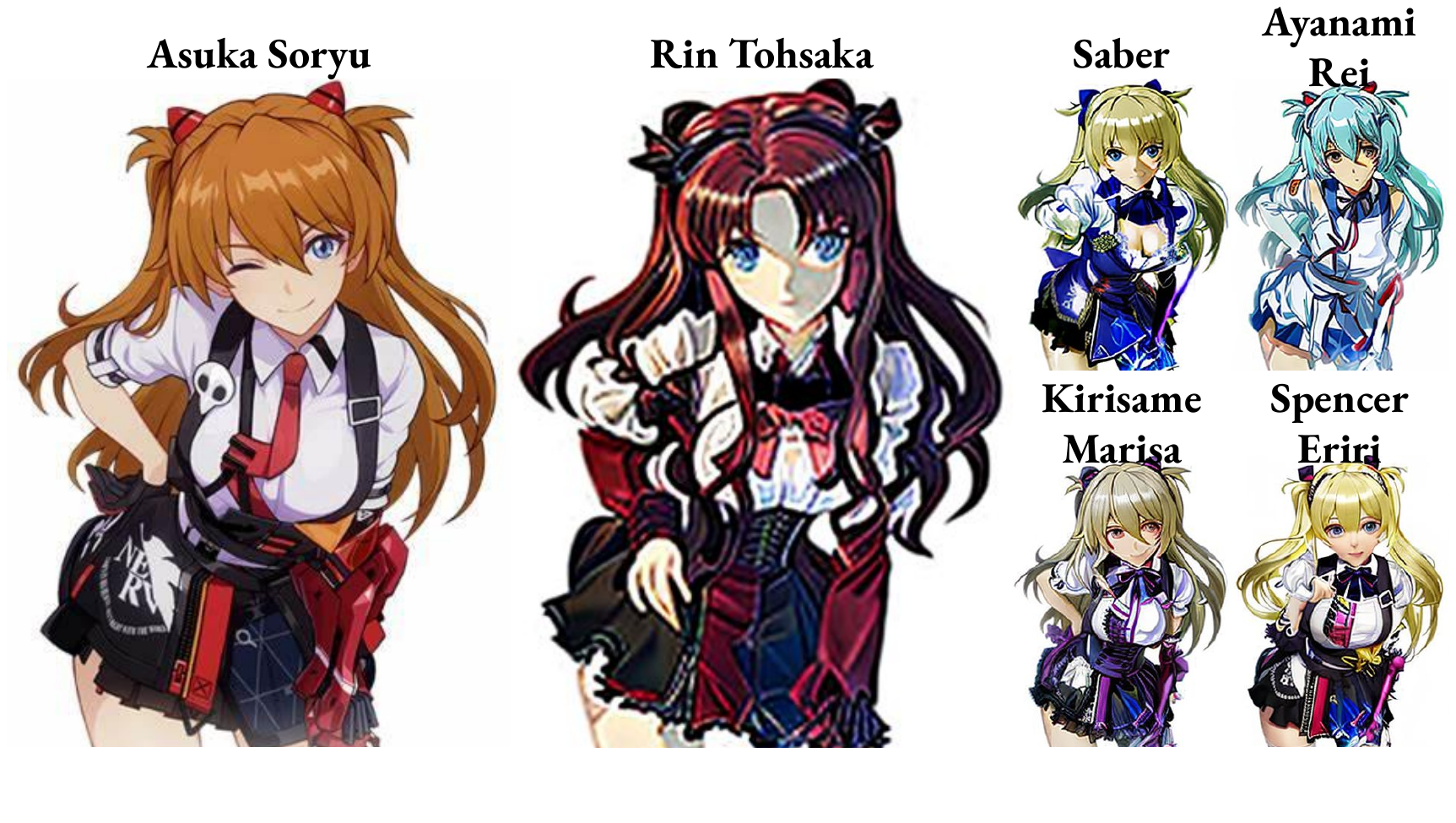}
    \vspace{-10pt}
    \caption{Additional results of InfEdit in multi-modal editing.} 
\end{figure*}

\clearpage

\begin{figure*}[!htp]
    \centering
    \includegraphics[width=0.6\linewidth]{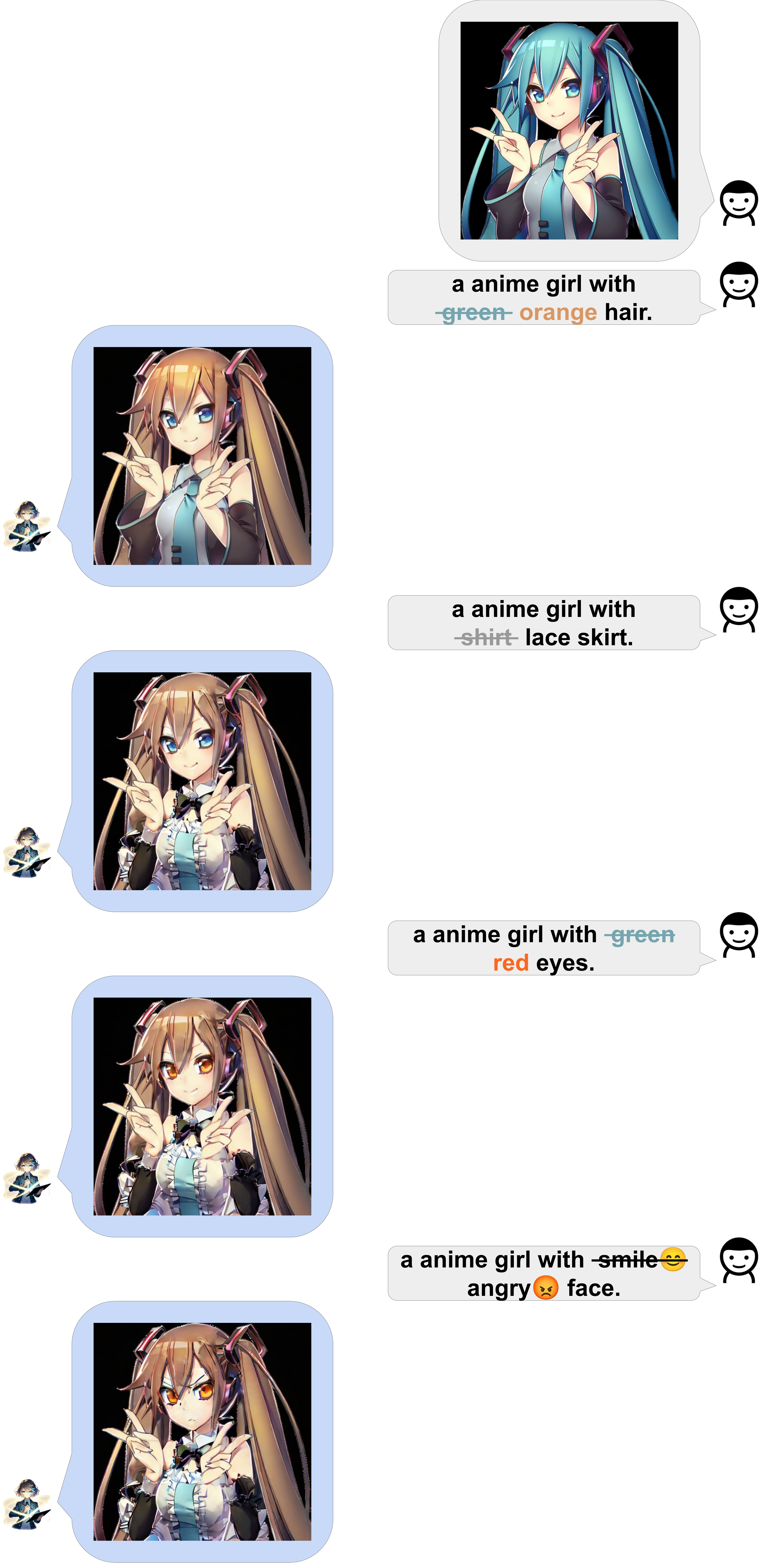}
    \caption{Multi-turn editing via InfEdit.} 
\end{figure*}

\clearpage

\end{document}